\documentclass[10pt,journal,compsoc]{IEEEtran}
\usepackage{times}
\usepackage{epsfig}
\usepackage{graphicx,subcaption}
\usepackage{amsmath}
\usepackage{amssymb}

\usepackage{color}
\usepackage{comment}
\usepackage{multirow}
\usepackage{pifont}
\newcommand{\cmark}{\ding{51}}%
\newcommand{\xmark}{\ding{55}}%
\usepackage{nccmath}
\usepackage{makecell}
\usepackage{afterpage}
\usepackage{booktabs}
\usepackage[font=small,labelfont=bf,justification=justified]{caption}
\usepackage{tabu}

\newif\ifdraft
\draftfalse
\drafttrue 

\definecolor{orange}{rgb}{1,0.5,0}
\definecolor{violet}{RGB}{70,0,170}
\definecolor{magenta}{RGB}{170,0,170}
\definecolor{dgreen}{RGB}{0,150,0}

\usepackage[normalem]{ulem}

\ifdraft
\newcommand{\PF}[1]{{\color{red}{\bf PF: #1}}}

\newcommand{\SH}[1]{{\color{blue}{\bf SH: #1}}}

\newcommand{\MS}[1]{{\color{dgreen}{\bf MS: #1}}}

\newcommand{\HR}[1]{{\color{orange}{\bf HR: #1}}}
\newcommand{\hr}[1]{{\color{orange} #1}}
\else
\newcommand{\PF}[1]{}

\newcommand{\SH}[1]{}

\newcommand{\MS}[1]{}

\newcommand{\HR}[1]{}
\newcommand{\hr}[1]{#1}\fi

\newcommand{\bB}{\mathbf{B}}

\newcommand{\bD}{\mathbf{D}}

\newcommand{\bI}{\mathbf{I}}

\newcommand{\bK}{\mathbf{K}}
\newcommand{\bR}{\mathbf{R}}
\newcommand{\bM}{\mathbf{M}}

\newcommand{\bT}{\mathbf{T}}

\newcommand{\TV}{{tv}}
\newcommand{\TI}{{ti}}

\newcommand{\mL}{\mathcal{L}}
\newcommand{\mF}{\mathcal{F}}
\newcommand{\mS}{\mathcal{S}}

\newcommand{\bm}{\mathbf{m}}
\newcommand{\bn}{\mathbf{n}}

\newcommand{\bq}{\mathbf{q}}

\newcommand{\bp}{\mathbf{p}}
\newcommand{\bv}{\mathbf{v}}
\newcommand{\bg}{\mathbf{g}}

\newcommand{\mD}{\mathcal{D}}
\newcommand{\mE}{\mathcal{E}}

\newcommand{\simab}{\text{sim}}

\usepackage{multirow}

\ifCLASSOPTIONcompsoc
  \usepackage[nocompress]{cite}
\else
  \usepackage{cite}
\fi

\ifCLASSINFOpdf
\else
\fi
\newcommand{\etal}{\textit{et al.}}

\begin{document}

\title{Temporal Representation Learning on Monocular Videos for 3D Human Pose Estimation}
%

\author{Sina Honari${^1}$\thanks{sina.honari@epfl.ch}, Victor Constantin${^1}$, Helge Rhodin${^2}$, Mathieu Salzmann${^1}$, Pascal Fua${^1}$\\
	{\normalsize ${^1}$CVLab, EPFL, Lausanne, Switzerland} \\
	{\normalsize ${^2}$Imager Lab, UBC, Vancouver, Canada}}

\IEEEtitleabstractindextext{%

\begin{abstract}
In this paper we propose an unsupervised feature extraction method to capture temporal information on monocular videos, where we detect and encode subject of interest in each frame and leverage contrastive self-supervised (CSS) learning to extract rich latent vectors. Instead of simply treating the latent features of nearby frames as positive pairs and those of temporally-distant ones as negative pairs as in other CSS approaches, we explicitly disentangle each latent vector into a time-variant component and a time-invariant one. We then show that applying contrastive loss only to the time-variant features and encouraging a gradual transition on them between nearby and away frames while also reconstructing the input, extract rich temporal features, well-suited for human pose estimation. Our approach reduces error by about 50\% compared to the standard CSS strategies, outperforms other unsupervised single-view methods and matches the performance of multi-view techniques. When 2D pose is available, our approach can extract even richer latent features and improve the 3D pose estimation accuracy, outperforming other state-of-the-art weakly supervised methods. 

\end{abstract}

}

\maketitle


\section{Introduction}

While supervised body pose estimation is rapidly becoming a mature field, the bottleneck remains the availability of sufficiently large training datasets containing in-the-wild images or depicting all kinds of human motions.
While some datasets  \cite{Ionescu14a, Mehta17a} provide relatively large-scale annotations in controlled environments, the set of motions is limited and does not capture the vast diversity of possible human poses, such as the ones observed in athletics or diving. An effective way to address this is to leverage unsupervised data to learn a low-dimensional representation of pose-related features. Then, it only takes very little annotated data to train a regressor to predict 3D poses from this representation. Key to the success is a good unsupervised learning objective, for which many existing techniques exploit the availability of multi-view footage~\cite{Pavlakos17,Tung17b,Zhou17a,Rhodin18a,Rhodin19a}. However, capturing data with multiple cameras increases complexity because it requires synchronizing the cameras and calibrating them to obtain the camera parameters that several of these models~\cite{Pavlakos17, Zhou17a, Rhodin18a,Rhodin19a} require during training.

In this paper, we introduce an alternative unsupervised representation learning strategy using videos acquired by a {\it single} RGB camera. To this end, we build on the idea of contrastive self-supervised (CSS) learning~\cite{Oord18,Hyvarinen16,Chen20a}. For any given sample, CSS aims to maximize the similarity to a positive sample and the dissimilarity to a negative one. In our context, one way to do so is to treat a  video frame close to the one of interest as positive, and a temporally-distant one as negative~\cite{Anand19,Sermanet18}. Unfortunately,  this simple strategy is insufficient because it does not account for the fact that, when someone moves, some features, such as garment appearance, remain constant over time, even as the person's pose changes. As a result, contrastive learning should be applied only to the features changing across time and not to all features.
To overcome this, we learn a representation that explicitly separates each latent vector into a {\it time-variant} component and a {\it time-invariant} one and apply the contrastive loss only to the {\it time-variant} component. 

\begin{figure*}[htbp]
	\centering
	\includegraphics[width=1\textwidth]{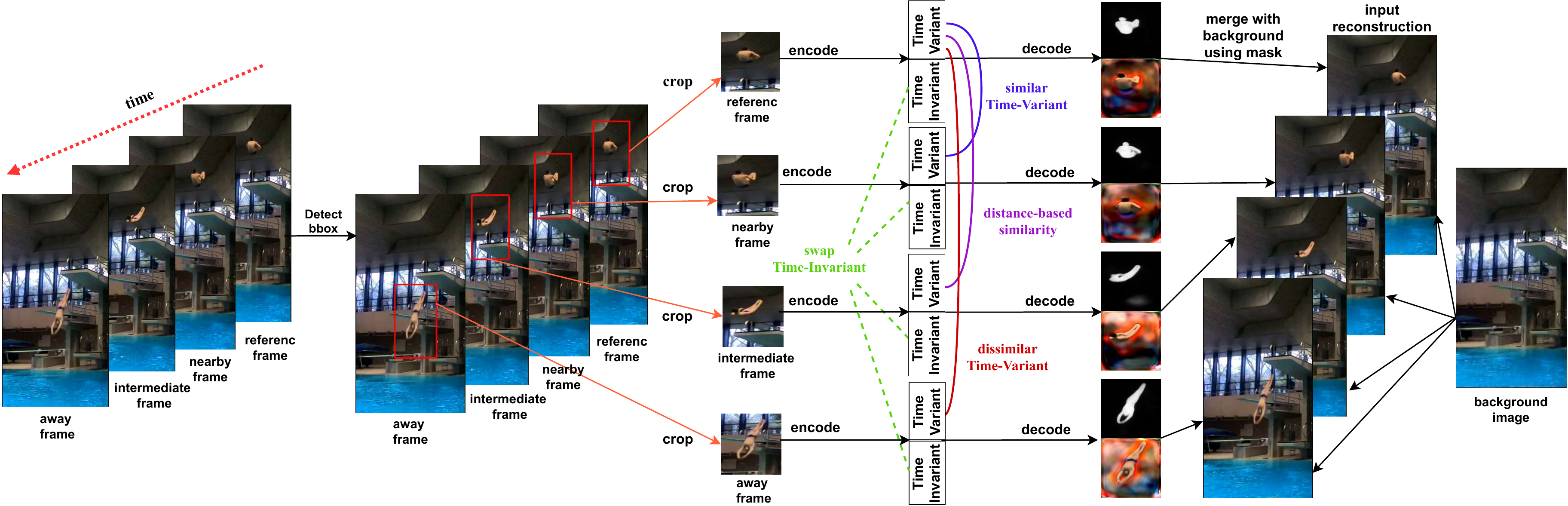}
	\caption{\small {\bf Approach.} Given a reference frame, we sample a temporally close, far, and intermediate---between close and far ---frames. The model first detects the person in all frames using a spatial transformer network. The detected regions are then cropped and encoded into \textit{time-variant} and \textit{time-invariant} components. The time-variant of the reference frame is made similar to that of the nearby frame and dissimilar to that of the away frame, while a distance-dependent similarity or dissimilarity is enforced between the reference and intermediate frames. To promote invariance of the time-invariant components, they are swapped with other frames in the video. The model then decodes each frame into a segmentation mask and an image. The decoded image is then merged with the background using the bounding box location and the decoded mask. No annotations are used (neither for detection, nor for segmentation) and the same network weights are used for all frames. At test time, the model can process a single frame.}
	\label{fig:arch}
\end{figure*}

Our approach, depicted in Fig.~\ref{fig:arch}, differs from typical CSS strategies in four ways: First, we apply the contrastive formulation only to the time-variant component, as opposed to the entire latent space. Second, we encourage a gradual transition of the time-variant component from temporally close to temporally distant frames. Third, we perform image re-synthesis by mapping the latent vectors back to images. Finally, when observing subjects in free fall such as divers, we incorporate a gravity-encoding  prior to help with the detection of the subject. Our experiments demonstrate that these four components help the model learn in an unsupervised manner a rich, low-dimensional representation of human motions from ordinary video sequences acquired using standard monocular RGB cameras. We also show when 2D labels are available, one can use them to extract richer time-variant features, such that they contain more relevant pose information.


We demonstrate the benefits of our approach for pose estimation on five real-world RGB datasets. It outperforms other unsupervised monocular approaches and yields results on par with unsupervised multi-view setups, while being much easier to deploy. In particular, it can exploit both monocular and multi-view videos when available, which a purely multi-view setup cannot, and can handle changing backgrounds. When using 2D labels, our approach outperforms other weakly supervised methods. The disentangled latent representation also lends itself for editing operations, such as pose and appearance transfer.



\section{Related Work}
\label{sec:related}

Unsupervised and self-supervised learning~\cite{Bengio13a,Doersch17} techniques have received considerable attention in recent years as a means to decrease the amount of annotated data required to train deep networks. We briefly review some of the methods most closely related to our work.

\subsection{Single-View Approaches}
Unsupervised learning for general video processing has been extensively researched. For example, LSTMs have been used to learn representations of video sequences~\cite{Srivastava15b}. Similarly, optical-flow and color diversity across frames have been leveraged to learn latent features corresponding to motion and appearance~\cite{Wang19l}.  Multiple papers \cite{Misra16,Lee17,Fernando17,Xu19} use the temporal order of frames to learn latent features from videos.  In particular, the models of~\cite{Misra16, Fernando17} classify whether a set of frames have correct ordering or not, whereas those of~\cite{Lee17, Xu19} predict the order of the frames or clips as a classification task. In~\cite{Wang20b}, instead of relying on temporal ordering, a model is trained to predict the pace of a video subsampled at different rates. The aforementioned work also leverages a CSS loss across clips, where positive pairs are clips from the same video and negative pairs are clips from different ones. None of these techniques, however, were designed for pose estimation. For example, predicting temporal order~\cite{Misra16,Lee17,Fernando17, Xu19} is best suited for activity recognition, for which only the order of the observed event truly matters.
Furthermore, the contrastive learning strategy used in~\cite{Wang20b} aims to extract features that are invariant across video frames. In contrast to these works, to predict a continuously changing body pose, we need to capture variations between the frames.

Several works have nonetheless focused on extracting time-varying features, yet in different contexts. Harley \etal \cite{Harley20} use RGB and depth from $n$ time-steps to map 2.5D features to a 3D latent space, and then use the camera pose at time-step $n+1$ to predict the 2D latent features for that time-step. 
While~\cite{Harley20} takes depth as input in addition to RGB, other works~\cite{Sermanet18, Anand19, Chen20a} leverage CSS, with~\cite{Sermanet18} and ~\cite{Anand19} using it for only single-view RGB videos. However, they do no use any decoder to reconstruct the images from the latent features, thus opening the door to losing valuable information about the observed context. In addition, these approaches do not split the latent space, which is equivalent to only learning the time-variant component.
We will demonstrate that for the pose estimation task our encoding/decoding approach with latent split that applies the contrastive loss only to the time-variant component yields richer features. 

Unsupervised learning methods have also been developed for 2D keypoint estimation~\cite{Thewlis17a,Zhang18b, Wiles18a,Jakab18, thewlis19, lorenz19, jakab20}. Some of these~\cite{Zhang18b, Wiles18a, Jakab18,  lorenz19, jakab20} learn separate features for pose and appearance, as we do. However, being designed for 2D keypoint estimation, they either explicitly detect 2D keypoints or generate heatmaps, and extending them to 3D pose estimation remains unadressed. Furthermore, whereas we learn to detect our subject without supervision, these methods rely on ground-truth bounding box locations.



\begin{figure*}[thbp]
	\centering
	\includegraphics[width=1\textwidth]{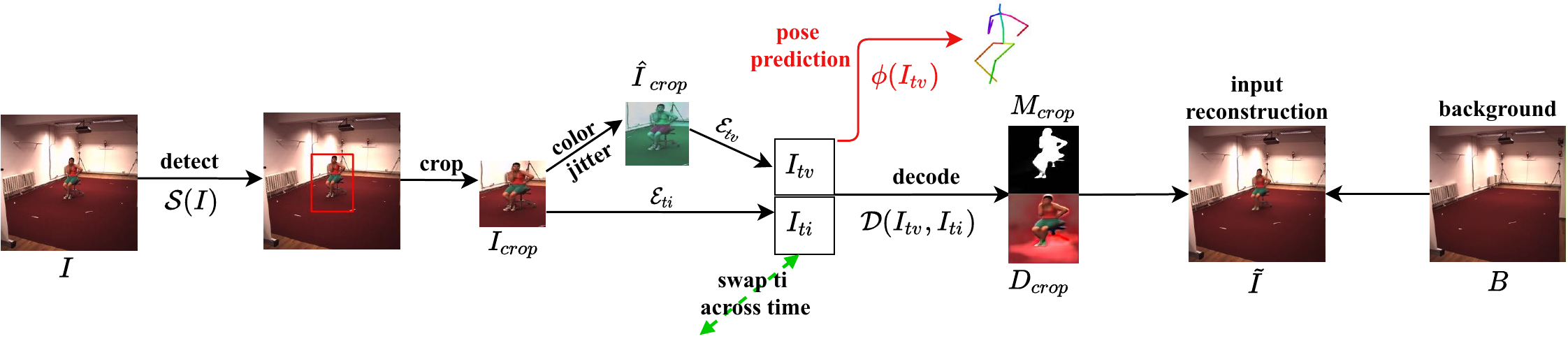}
	\caption{\small \textbf{Model architecture.} Using spatial transformer network (STN) $\mS$, a bounding box is detected and cropped on an input image $\bI$. The cropped image, $\bI_\text{crop}$, is then passed to an encoder $\mE_{ti}$ that predicts time-invariant $\bI_\TI$ component. This component is swapped across time with other frames to enforce disentanglement. A color jittered variant of the crop $\hat{\bI}_\text{crop}$ is passed to another encoder $\mE_{tv}$ to predict time-variant $\bI_\TV$ features. Both encoders share parameters, except in the output layer. The combined features are then passed to a decoder $\mD$ that outputs a mask $\bM_{\text{crop}}$ and RGB crop $\bD_{\text{crop}}$. These two outputs are then put to the image resolution using inverse STN operation and then merged with background $\bB$ to output the reconstruct image $\tilde{\bI}$. This constitutes the self-supervised model, which is then frozen. In the second phase (shown in red), predicted $\bI_\TV$ is passed to a shallow network $\Phi$ to predict the 3D pose.
	}
	\label{fig:diag}
\end{figure*}

\subsection{Multi-View Approaches}

The difficulty of performing unsupervised learning using a single camera has led to the development of several multi-view approaches~\cite{Zhou17a, Moniz18,Sermanet18, Tian19, Rhodin18b,Rhodin19a}. 
While some of them are semi-supervised~\cite{Rhodin18a, Mitra20}, here, we focus only on fully unsupervised learning, where no pose labels are used when training the feature extraction model. 

Among the unsupervised methods, the strategy of~\cite{Rhodin18b, Rhodin19a} is closest in spirit to ours. It uses a multi-camera setup to learn a low-dimensional latent vector that disentangles pose from appearance. An effective mapping from this latent space to 3D poses can then be trained using only a small amount of annotated data. In contrast to this approach, we aim to learn a latent representation from videos acquired by a \emph{single} camera. To do so, we leverage contrastive learning and propose multiple innovations compared to the standard CSS approaches.

\subsection{Disentanglement}

The goal of disentanglement is to find independent and interpretable factors of variation \cite{Kim18, Chen18f, Karras19, Chen16g, Wang20g, Remelli20a} that makes the data easier to process or understand. For example, \cite{Worrall17} encodes facial images into disentangled components such as illumination, head yaw, and pitch angles. On the other hand, \cite{Remelli20a} disentangles encoded image features from camera's view point to a canonical representation that can be more easily processed by the model for 3D pose estimation. In \cite{Wang20g} it has been used to separate appearance and motion in a generative model. One closely related approach is DrNet~\cite{Denton17}, which disentangles each frame into pose and content. It uses a discriminator that distinguishes pose component extracted from the same video from the ones extracted from different videos and encourages the encoder to produce latent representations that do not carry any information about whether they come from the same or different videos. This, however, fails in cases where the two videos feature different kinds of activities---for example, running in one and sitting on the floor in the other. In this scenario, the pose still indicates if the frames come from the same video or not, which precludes disentanglement. 

Another closely related approach is ~\cite{Rhodin18a} that use a multi-view setup to disentangle pose from appearance, by rotating the pose from one view to another while swapping the appearance of a subject with other frames of the same subject in the dataset. Contrary to these approaches, we split features on monocular videos into time-variant and time-invariant components, by using contrastive learning to ensure time-variant contains only temporal features that change across time, while also making sure it does not capture identity information, such as appearance, through input jittering to time-variant and swapping of time-invariant across time.
\section{Method}
\label{sec:method}

Our goal is to extract rich latent features,  without any supervision, from single-camera videos featuring people. We aim to capture information only about the foreground subjects and to leverage the similarity or dissimilarity of frames given their distance in time. A key observation is that global object appearance tends to remain constant over time while pose usually changes from frame to frame, which is something our latent vectors should reflect. 


Encoding the entire image is expected to yield inferior performance because the resulting latent variables would also capture irrelevant background information. To address this, we guide the model to detect the subject of interest in an unsupervised manner, using image reconstruction. We then only encode features of the foreground subject, to which we apply the contrastive loss. 


We train an encoder that splits latent features into \textit{time-variant} and \textit{time-invariant} components. The latter capture features that remain consistent over time, such as a person's clothing and appearance, while the former models the time-changing elements in each frame, such as body pose. We then apply the contrastive loss only to the time-variant elements.
 Without this split, the latent components learned using CSS would struggle to reliably differentiate temporally-distant frames, because they would still share information, such as identity.
Finally, we take time-invariant and time-variant components to re-synthesize the input.
 We show in the experiments that decoding images help the model learn much richer features, as the model has to capture all relevant information on the subject to reconstruct the input. Fig.~\ref{fig:arch} summarizes our approach.

\subsection{Model Architecture and Training}
\label{sec:arch}

We tackle real-world applications where the foreground subject is small and occupies only a fraction of the image. To this end, we use a neural network architecture with an attention mechanism to separate foreground from background. As shown in Fig.~\ref{fig:diag}, given an input image $\bI$, a spatial transformer network (STN)~\cite{Jaderberg15}  $\mS$ extracts four parameters that define the subject's bounding box, two non-homogeneous scales $s^x, s^y$ and its center coordinates $u^x, u^y$, yielding $\mS(\bI)=(s^x, s^y, u^x, u^y)$. These parameters are used to crop an image patch $\bI_\text{crop}$. A color-jittered version of the crop $\hat{\bI}_\text{crop}$ is passed to the time-variant encoder $\mE_{tv}$ such that $\mE_{tv}(\hat{\bI}_\text{crop}) = \bI_\TV$, with $\bI_\TV$ the time-variant latent component. A crop of another frame ${\bI}^{'}_\text{crop}$ of the same video is passed to a time-invariant encoder $\mE_{ti}$ such that $\mE_{ti}({\bI}^{'}_\text{crop}) = \bI^{'}_\TI$, where $\bI^{'}_\TI$ is the time-invariant component of the latent vector for image $\bI^{'}$.  
Both encoders share parameters except at the output layer. A decoder $\mD$ then takes $\bI_\TV$ and $\bI^{'}_\TI$ as inputs and decodes them into a mask $\bM_{\text{crop}}$ and an RGB crop $\bD_{\text{crop}}$. The mask and RGB crop are then used to re-synthesize a full-resolution image by using the inverse transformation of the STN and merging the result with a background image $\bB$. This operation can be written as 
\begin{align}
\tilde{\bI}=   \bM \times \bD + (1- \bM) \times \bB \;,
\label{eq:synthesize}
\end{align}
where $\tilde{\bI}$ is the reconstructed image. Fig.~\ref{fig:diag} depicts this process. Note that no labels are used here, neither for detection, nor for segmentation.
We train the networks $\mS$, $\mE$, and $\mD$ such that they jointly accomplish three goals:
\begin{itemize}
 \item Encoding information that relates only to the foreground object (Section \ref{sec:reconst});

 \item Disentangling the time-variant and time-invariant components (Section \ref{sec:disentanglement});
 
 \item Encoding time-variant components such that they are more similar for poses that are close in time than for temporally distant ones (Section \ref{sec:CSS}).

 
\end{itemize} 
%

\subsubsection{Reconstruction Loss}
\label{sec:reconst}

To compare the reconstructed image from Eq.~(\ref{eq:synthesize}) with the original one, we use the L2 loss on the image pixels as well as on the features extracted from layers 1 to 3 of the ResNet18~\cite{He16a} model. This can be written as
\begin{align}
\mL_{\text{reconst}} =  \lambda{\| \bI - \tilde{\bI}\|}_2^2 + \rho \sum_{l=1}^{3}{\| \text{Res}_{l}(\bI) - \text{Res}_{l}(\tilde{\bI})\|}_2^2 \; .
\label{eq:reconst}
\end{align}
As discussed above, we use a background image when re-synthizing the input image, hence this reconstruction loss encourages our latent representation to focus on the foreground object.

\subsubsection{Contrastive Loss} 
\label{sec:CSS}

To encourage time-variant components that are close in time to be close to each other and those that are far in time not to be, we rely on a contrastive loss. However, we diverge from standard CSS formulations~\cite{Oord18, Anand19, Sermanet18,Chen20a} in the two following aspects: 
1) Instead of applying the contrastive loss to the entire latent space, we enforce it only on the time-variant features, as only part of the latent features should evolve over time. 2) We enforce a gradual transition of the latent representation from nearby frames to away frames. Below, we discuss this in more detail, starting from the standard CSS formulation.

Given a reference input $\bI$, a positive sample $\bI^{+}$, and $n-1$ negative samples $\bI_k^-$ for $1 \leq k < n$, standard CSS approaches~\cite{Chen20a, Oord18, Anand19} minimize the loss
\begin{align}
\mL_{\text{CSS}} =  - \log \frac{\text{exp}(\simab(\mF(\bI), \mF(\bI^{+}))/ \tau)}{\sum_{k=1}^{n} \text{exp}(\simab(\mF(\bI), \mF(\bI_k^{-/+})) / \tau)} \; , 
\label{eq:cssLoss}
\end{align}
where $\bI_k^{-/+}$ indicates a set of one positive and $n-1$ negative examples. $\mF(\bI)$ represents the features encoded by the network $\mF$ given frame $\bI$ as input and $\simab(\bm, \bn) = ({\bm^{T}\bn})/({\|\bm\|  \|\bn\|})$. In~\cite{Sermanet18}, the contrastive loss is defined instead as a triplet loss 
\begin{align}
\mL_{\text{CSS}} \!=\!  \left[ \| \mF(\bI) \!-\! \mF(\bI^{+}) \|^2 + \beta \right] \!< \!\| \mF(\bI) \! -  \! \mF(\bI^{-}) \|^2 \; ,
\label{eq:triplet}
\end{align}
with $\beta > 0$, meaning that the squared L2 distance between a positive pair should be smaller than that between a negative one by at least $\beta$.  In any event, neither definitions of $\mL_{\text{CSS}}$ above define the loss relative to the temporal distance of the frames. This is what we change by introducing a distance-based similarity loss. 
%
%

\vspace{.3cm}
\noindent\textbf{Distance-based Similarity Loss (DSL).} 
Our goal is to compare every pair of samples directly and to define a loss that enforces similarity or dissimilarity between any two samples depending on their temporal proximity. 

\begin{figure}[htbp]
	\centering
	\includegraphics[width=0.45\textwidth]{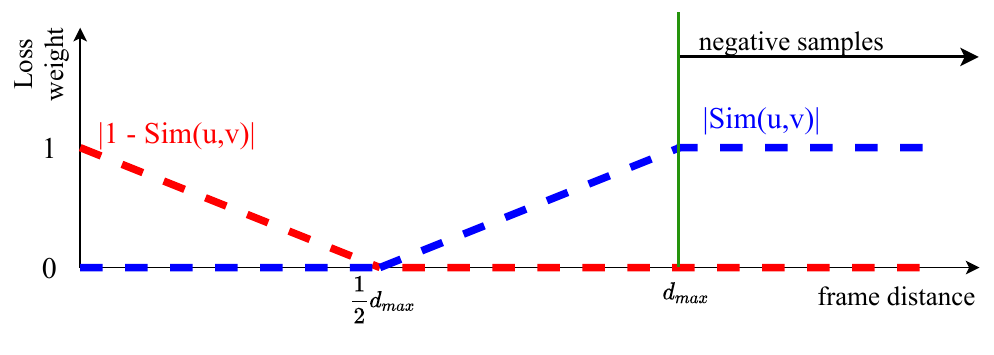}
	\caption{\small {\bf Distance-based Similarity Loss (DSL).} The red line shows the \textbf{weight}, $ (1- \frac{2d}{d_\text{max}})$,  applied to $|\simab(\bm, \bn) - 1 |$ when two frames are close (similarity measure) and the blue line shows the \textbf{weight}, $\min(\frac{2d}{d_\text{max}} -1, 1)$, applied to $|\simab(\bm, \bn)|$ when the frames are away (dissimilarity measure). Both weights depend on the temporal distance between the frames. As the frame distance increases, the similarity weight is reduced, eventually going to zero at $d_\text{max}/2$, and the dissimilarity weight increases, eventually reaching one at $d_\text{max}$, the frame from which negative samples are taken.}
	\label{fig:loss}
\end{figure}

Let $\bI_{tv}^m$ and $\bI_{tv}^n$ be the time-variant component of the vectors extracted from two images captured at times $t_m$ and $t_n$. If $|t_m-t_n|$ is small, we would like $\simab(\bm, \bn)$, where $\simab()$ is the similarity function used in Eq.~(\ref{eq:cssLoss}), to be close to one. By contrast, if $|t_m-t_n|$ is larger than a threshold distance $d_\text{max}$, we would like $\simab(\bm, \bn)$ to be zero. 
%
%
To this end, we introduce the loss function
%
%
%
\begin{align}
\label{eq:DSL}
&\mL_{\text{DSL}}(\bI_{tv}^m, \bI_{tv}^n) =  \\
& \begin{cases}
(1- \frac{2d}{d_\text{max}}) \; |1 -\simab(\bI_{tv}^m, \bI_{tv}^n) |, \; \; \; \; & \text{if} \; 0 \le d \le  d_\text{max}/2 \; ,\nonumber \\
\min(\frac{2d}{d_\text{max}} -1, 1) \; |\simab(\bI_{tv}^m, \bI_{tv}^n)|,  & \; \text{otherwise,}
\end{cases}
\end{align}
where $d=|t_m-t_n|$ is the temporal distance between the two frames. As shown in Fig.~\ref{fig:loss},  $\mL_{\text{DSL}}$ is piecewise linear in terms of $\simab(\bI_{tv}^m, \bI_{tv}^n)$. It favors similarity for $0 \le d \le  d_\text{max}/2$, and dissimilarity for $d > d_\text{max}/2$, with similarity decreasing and dissimilarity increasing as $d$ increases.

In practice, to learn time-variant components in our framework, for any given reference input frame $\bI^r$, we sample three other frames, a nearby one $\bI^{n}$ (positive sample), a far away one $\bI^{a}$ (negative sample taken at least at $d_\text{max}$), and an intermediate one, $\bI^{in}$ (acquired after $\bI^{n}$ and before $\bI^{a}$). We then compute 
\begin{small}%
\begin{align}%
 \mL_{\text{DSL}}^{\text{all}}  \! =  \! \mL_{\text{DSL}}(\bI^r_{\TV},\bI^{n}_{\TV})  \! +  \! \mL_{\text{DSL}}(\bI^r_{\TV},\bI^{a}_{\TV})  \! +  \! \mL_{\text{DSL}}(\bI^r_{\TV},\bI^{in}_{\TV}) \; .
 \label{eq:fourImageLoss}
\end{align}%
\end{small}%
The above loss allows us to compare the reference frame with similar, dissimilar, and intermediate frames directly.

 
\subsubsection{Disentanglement}
\label{sec:disentanglement}

To force the network to disentangle the latent into time-variant and time-invariant components, we apply three things: 1) we swap the time-invariant ones of different frames of the same video during training. This prevents the time-invariant components from encoding information about the subject's pose, 2) we use a color-jittered variant of the cropped image $\hat{\bI}_\text{crop}$ to predict the time-variant features $\bI_\TV$. This forces $\mE_{tv}$ to focus on the pose and not the texture when predicting $ \bI_\TV$, and 3) we use contrastive loss to make temporally distant time-invariant dissimilar.
Note that points 1 and 3 enforce the time-invariant component to capture the identity information, as we will later show in Section \ref{sec:exp_disentangle}. Also note that reconstruction is needed to make points 1 and 2 impactful, as otherwise the model does not encode relevant information.


\begin{figure*}[htbp]
	\centering
	\includegraphics[width=0.6\textwidth]{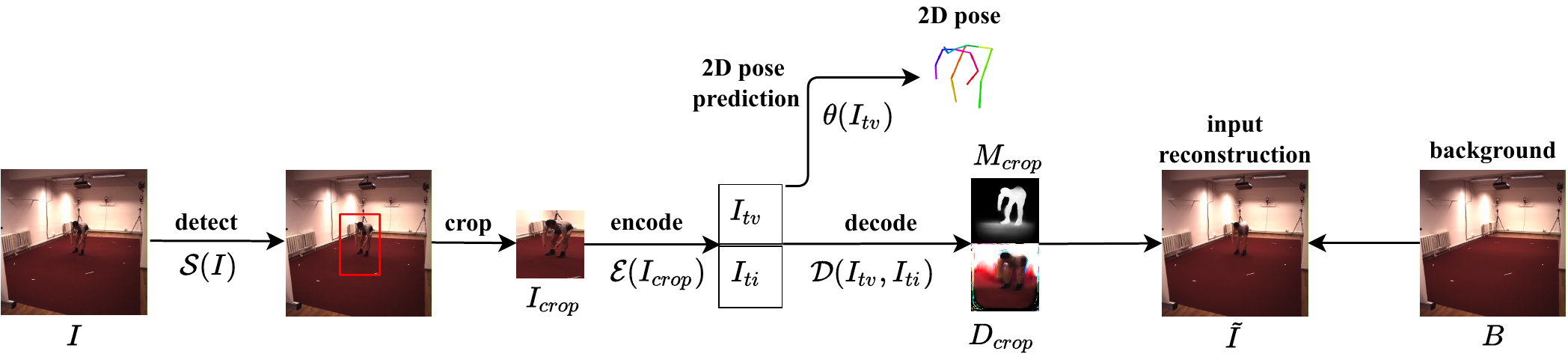}
	\hspace{0.5cm} \vline \hspace{0.5cm}
	\includegraphics[width=0.2\textwidth]{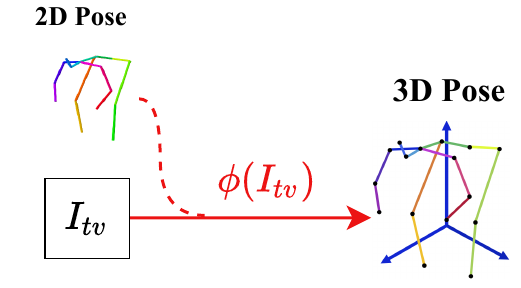}
		\caption{\small {\bf Using Weak Labels.} 	 
			(left)When 2D labels are available in the self-supervised phase, we apply 2D regression on time-variant features in addition to image reconstruction. (right) In the 2nd phase,  $\bI_{\TV}$ features and optionally 2D pose can be used to predict the 3D pose.
	}
	\label{fig:weak_labels}
\end{figure*}

\subsection{Gravity as a Supervisory Signal}
\label{sec:gravity}

The work reported here was initially motivated by a project that involves modeling competitive divers. Because the divers are in free fall, we can leverage Newtonian mechanics as follows. The diver's center of gravity (CG) is subject to gravitational acceleration. Ignoring the air resistance that remains small until the diver reaches the water, $\bp(t)$, the CG location at time $t$, can be written as $\bp(t)=0.5 \bg t^2 + \bv_0 t+ \bp_0$,  where $\bg$ is the gravitational vector, $\bv_0$ the CG initial speed, and $\bp_0$ its initial location. We know neither $\bv_0$ and $\bp_0$ nor when the jump started. Nevertheless, we can leverage the fact that the acceleration, that is, the second derivative of $\bp$ is constant and equal to $\bg$. To this end, we only need to consider four {\it equidistant} times $t_1$, $t_2$, $t_3$, $t_4$. We can then approximate the acceleration at $t_2$ and $t_3$ using finite differences and write that they should be equal. This yields
\begin{align}
	2 \bp(t_2) - \bp(t_1) - \bp(t_3) &= 2 \bp(t_3) - \bp(t_2) - \bp(t_4) \nonumber \\
	\Rightarrow 3 \bp(t_2) - 3 \bp(t_3) &+ \bp(t_4)  - \bp(t_1) = 0 \; . \label{eq:gravity}
\end{align}
In practice, two difficulties arise in imposing this simple constraint. First, it is not obvious where someone's CG is given that it depends on their posture and on the weight of each body part. Second, we only have access to 2D projections of $\bp(t)$ in the image plane and not the 3D locations.  

Recall from Section~\ref{sec:arch} that our model first applies an STN ${\cal S}$, which, for each frame, outputs a bounding box around the subject parameterized by  two scales and two translations,  $s^x, s^y, u^x, u^y$. We address the first problem by approximating the projection of $\bp(t)$ as the center of the bounding box. To address the second problem, we consider that the diver's CG travels in a plane. If it were parallel to the image plane, the value of $u^y$
would vary linearly with the CG height and the relationship of Eq.~(\ref{eq:gravity}) would hold exactly, implying that
\begin{align}
	3 u^y_2 - 3 u^y_3 &+ u^y_4  - u^y_1 = 0 \; , \label{eq:gravityProj}
\end{align}
where $u^y_t$ denotes the value of $u^y$ at time $t$. Even when the diver does not travel in a plane parallel to the image one, the frame-to-frame change in distance to the camera of the CG typically remains small  and perspective effects are negligible with respect to the vertical motion. Eq.~(\ref{eq:gravityProj}) is therefore a good approximation, which we will show in Section~\ref{sec:sup-gravity} of the supplementary material. We therefore impose the constraint of Eq.~(\ref{eq:gravityProj}) as a soft constraint by adding a new loss term into our training procedure. 

As shown in the supplementary material, displacement along the $x$ axis occurs is much smaller than  that along the $y$ axis, which we therefore focus on.  Hence, we take the new loss term to be
\begin{align}
	\mL_{\text{const-acc}} =\left\| \left( u^y_1 + 3 \, u^y_3 \right) - \left( u^y_4 + 3 \, u^y_2 \right) \right\|^2.
\end{align}
%
This definition does not differentiate between moving up or down, whereas, in practice, we know that the diver goes down. In our implementation, this corresponds to $u^y_t$ decreasing. Hence, we also introduce an {\it order loss} $\mL_{\text{order}}$ that we define as
\begin{align}
	\mL_{\text{order}} &= \sum_{t=1}^3  \max \left( 0, \tau - \left(  u^y_t - u^y_{t+1} \right) \right) \; ,
\end{align}
where $\tau$ is a threshold to satisfy a minimum distance between frames $t$ and $t+1$. 

In free-fall, the detector could still make the bounding box bigger to compensate for the aforementioned losses. We therefore prevent the scales $s_x$ and $s_y$  of the bounding boxes from varying abruptly from frame to frame by defining a {\it scale loss} $\mL_{\text{sc}}$ as
\begin{align}
	\mL_{\text{sc}} =  \sum_{t=1}^3   \left\| (s^x_t, s^y_{t}) - (s^x_{t+1}, s^y_{t+1}) \right\|^2.
\end{align}
Putting this all together, the tracking loss is then equal to
\begin{align}
	\mL_{\text{track}} &= \mL_{\text{const-acc}} + \mL_{\text{order}}  + \mL_{\text{sc}} \;.
	\label{eq:track}
\end{align}

\subsection{Leveraging Weak Labels}
\label{sec:weak_labels}

When 2D pose labels are available, we can use them to learn richer latent features. In particular, as shown in Fig. \ref{fig:weak_labels}, we apply a multi-layer-perceptron network $\Theta$ only on the time-variant feature $\bI_{\TV}$ and regress the 2D pose $\bq^{k}_{2D}$, yielding

\begin{align}
	\mL_{\text{pose}}^{2D} = \frac{1}{N}\sum_{k=1}^{N} \| \Theta(\bI_{\TV}^{k}) - \bq^{k}_{2D}\|^{2} \; .
	\label{eq:pose_2D}
\end{align}

\subsection{Training and Inference}
\label{sec:training}

To train our network, we minimize the total training loss 
\begin{align}
\mL_{\text{total}}  &=\mL_{\text{reconst}} + \alpha \mL_{\text{DSL}}^{\text{all}}  + \gamma \mL_{\text{track}} + \eta \mL_{\text{pose}}^{2D}\;,
\label{eq:totalLoss}
\end{align}
where $\mL_{\text{reconst}}$ of Eq.~(\ref{eq:reconst}) favors a good image reconstruction, $\mL_{\text{DSL}}^{\text{all}}$ of Eq.~(\ref{eq:fourImageLoss}) applies a contrastive loss, and $\mL_{\text{track}}$ of Eq.~(\ref{eq:track}) is used when the subject is in free fall. $\mL_{\text{pose}}^{2D}$ is applied only when 2D pose is available. We carry out experiment with and without using 2D poses . During the minimization, we permute randomly the time-invariant components of the latent vectors to promote disentanglement as described in Section~\ref{sec:disentanglement}. This part of the training is unsupervised or weakly-supervised (depending on the usage of 2D labels). 

Once the self-supervised model is trained, its parameters are frozen and we use a small number of images with corresponding ground-truth poses for supervised training of a regressor $\Phi$ designed to associate a pose vector to the time-variant latent vector $\bI_{\TV}$ that the encoder ${\cal E}_{tv}$ has extracted from image $\bI_\text{crop}$. To this end, we adjust the weights of $\Phi$ by minimizing 
\begin{align}
\mL_{\text{pose}}^{3D} = \frac{1}{N}\sum_{k=1}^{N} \|\Phi(\bI_{\TV}^{k},  \hat{\bq}^{k}_{2D} ) - \bq^{k}_{3D}\|^{2} \; ,
\label{eq:pose}
\end{align}
where $\bI_{\TV}^{k}$ is the time-variant vector associated to the $k^{th}$ image and $\bq^{k}_{3D}$ the corresponding ground-truth 3D pose, in camera coordinates and pelvis centered, represented by a concatenation of the Cartesian coordinates of body joints. $\hat{\bq}^{k}_{2D}$ is the predicted 2D pose by the model, used only when 2D is leveraged in the initial self-supervised learning phase of the model.


\section{Results}
In this section, we present the datasets and baselines models used in our experiments, followed by quantitative results, ablation studies and analyses of the proposed approach.

\vspace{-0.2cm}
\subsection{Datasets.}
We evaluate our model on the following five datasets:
\vspace{-0.2cm}

\subsubsection{Human3.6M \cite{Ionescu14a} (H36M)}

As in~\cite{Rhodin19a}, we use subjects 1, 5, 6, 7, and 8 to train the self-supervised models and sub-sample every 5 frames of the training set, which yields 308,760 frames. We test our model on two different test sets:

\begin{itemize}
	\item Following ~\cite{Rhodin19a}, we use the PoseTrack2018 challenge~\cite{poseTrack18} setup to evaluate the accuracy of 3D human pose estimation. It involves 35,832 training images with 3D pose labels and 19,312 test samples. 
	
	\item When we use weak 2D labels, we report results on subjects 9 and 11, sub-sampled every 10 frames. As this test-set is used more often, it helps comparison with other weakly supervised methods.
\end{itemize}

\subsubsection{MPI-INF-3DHP  \cite{Mehta17a} (MPI)}
On this dataset, we consider two settings:
\begin{itemize}

	\item \textbf{MV-data:} The self-supervised models are trained using subjects S1 to S8 captured using multiple cameras. Since the background of the test-videos is different from that of the training ones, we train the self-supervised models on images with random backgrounds to help with generalization. To this end, we exploit  the foreground masks from the dataset to put the subject in front of  random images from the Internet. Note that these masks are used only to synthesize new training images but not by the networks themselves.
	
	\item \textbf{All-data:} Since self-supervised models do not require labels, we can leverage all the data to train them, whether single or multi-view. This corresponds to the transductive learning setting, and the self-supervised models are trained using S1 to S8 (multi-view) and TS1 to TS4 (single-view).
	
\end{itemize}
The video sequences are subsampled every 5 frames, yielding 198,096 training frames for MV-data and 202,905 for all-data. In both cases, we use subjects S1, S2, S3, S4, S5, and S8  to train the pose model, and S6 and S7 to form the validation set. For testing, we use TS1 to TS4. Because there are many pose labels, we use at most 2\% of the pose data for training, that is, 17,128  images. The test set comprises 2207 frames.

\subsubsection{Diving}

We acquired video sequences featuring one male and one female competitive diver. The dataset features dives from heights of 3, 5, 7.5, and 10 meters, most captured using multiple cameras and some using a single camera. We use both subjects to train the self-supervised models. As in the MPI case, we consider two settings: 
\begin{itemize}

	\item \textbf{MV-data:} Only multi-view data containing 16,469 frames.
	
	\item \textbf{All-data:} Both single- and multi-view data, yielding 19,464 training frames. 
\end{itemize}
The 3D pose is obtained by first annotating 2D pose in each view and then using triangulation on all views. This makes the 3D pose labels prone to noise, due to both self-occlusion in some views, and also mis-annotation by labelers, which contribute to the difficulty of 3D pose estimation on this dataset.


To train the pose model, we apply two settings:
\begin{itemize}
	
	\item \textbf{Diver-split:}  We use the female subject for training---splitting the videos into training and validation sets---and the male subject for testing. This setup yields up to 2,419 frames for supervised training, 2,288 frames for validation, and 2,112 frames for testing. This is the default setting. 
	
	\item \textbf{Dive-split:} Due to the strong bias of training on only one diver, we also split the dataset by dives, using unique and distinct dives in each split. While diver-split can better show generalization capability, it does not allow for proper evaluation due to the lack of identity diversity in the training set. The dive-split setup instead allows us to analyze how well the model can generalize to new dives, while not suffering from identity-related limitations.  This setup yields 2,789 training, 1,278 validation, and 2,724 testing frames. 

\end{itemize}

\vspace{-0.2cm}
\subsubsection{Skiing}

To evaluate our model on a dataset with moving cameras we use a Ski dataset~\cite{Ostrek19} that features 6 professional skiers, each going down a slalom course twice. The data is captured with two cameras. We use Subjects 1 to 5 for training and Subject 6 to evaluate the models. This yields 3,124 images for training and 636 images for testing for both self-supervised and pose models.

\subsubsection{3DPW}
We use this dataset for evaluation on in-the-wild images. Since our approach is designed for single subjects we only use single-subject videos, yielding 12,265 training and 14,803 testing frames.

\subsection{Baselines}

We compare our model to other unsupervised feature extraction methods, which are
\begin{itemize}

	\item \textbf{CSS~\cite{Anand19,Chen20a}:} Single-view CSS loss using Eq.~(\ref{eq:cssLoss}), without detection, image reconstruction, splitting the latent space into $\bI_{\TV}$ and $\bI_{\TI}$ components, or the DSL loss of Eq.~(\ref{eq:DSL}).
	
	\item \textbf{MV-CSS~\cite{Sermanet18}:} Multi-view CSS loss using the triplet loss of Eq.~(\ref{eq:triplet}), also without detection, reconstruction, splitting, or DSL loss.
	
	\item \textbf{DrNet~\cite{Denton17}:} Single-view approach using both reconstruction and latent splitting. It uses a discriminator for latent splitting.
	
	\item \textbf{NSD~\cite{Rhodin19a}:} Multi-view approach that performs both reconstruction and latent vector splitting.
	
\end{itemize}
We also evaluate variants of our own approach:
\begin{itemize}
		
	\item \textbf{DSL:}. This variant uses the contrastive loss of Eq.~(\ref{eq:DSL}) without detection, reconstruction, or decoding.
	
	\item \textbf{DSL-STN:}. This variant adds detection to the DSL model.
	
	\item \textbf{DSL-STN-Dec:}. This variant adds decoding to the DSL-STN model.
	
	\item \textbf{DSL-STN-Dec-Split (Ours):} This is our full approach that adds latent splitting to DSL-Dec-STN. It uses detection together with image-reconstruction and latent splitting. It applies Eq.~(\ref{eq:DSL}) for the contrastive loss.
	
	\item \textbf{AE-STN-Dec:}. This variant uses detection together with an auto-encoding image-reconstruction, but without contrastive training or latent splitting.
	
	\item \textbf{CSS-STN-Dec-Split:} This variant uses detection together with image-reconstruction and latent splitting, similar to Ours. However, it uses Eq.~(\ref{eq:cssLoss}) instead of Eq.~(\ref{eq:DSL}) for the contrastive loss, which does not leverage the DSL formulation.
	
\end{itemize} 
For a fair comparison, we followed the training protocol of~\cite{Rhodin19a} and used the same detector, encoder, and decoder for all models including the baselines.
In other words, unless explicitly specified, all these models are trained using the same architecture and model capacity, so that only the training losses differentiate them. For MV-CSS, we follow the recommendation of~\cite{Sermanet18} to sample one positive sample from another view at the same timestamp and one negative sample from a distant frame in the same video. For CSS and CSS-STN-Dec-Split, we use one positive sample from a nearby frame and two negative samples from away frames from the same video. For DSL-STN, we replace one of the away frames with an intermediate frame, sampled between nearby and away frames. We follow the procedures of~\cite{Denton17} and ~\cite{Rhodin19a} to train respectively DrNet and NSD. 

\subsection{Comparison against Unsupervised Methods} 
\label{sec:unsup_compare}

We compare our results to other unsupervised feature-extraction models on five datasets and if not otherwise specified report the N-MPJPE as a function of the percentage of labeled pose samples being used.

\subsubsection{H36M Dataset.}
In Table~\ref{tab:h36} we compare against both multi- and single-view approaches. We consistently outperform the other single-view methods and come close to NSD, even though it uses multiple views. Note that the contrastive approaches without decoding, whether single- or multi-view perform poorly on this task, which confirms that decoding helps to learn richer features. This further highlights the importance of using decoding together with contrastive learning and detection, especially when the subject occupies only a small portion of the image and small details should be captured for accurate pose estimation.


\begin{table}[ht]
	\centering
	\resizebox{1\linewidth}{!}{
		\begin{tabular}{|c|l|c|ccccc|}
			\hline
			& \multirow{2}{*}{\textbf{Model}}& \multirow{2}{*}{\textbf{Decodes?}} & \multicolumn{5}{c|}{\textbf{Percentage of Labeled Pose Data}} \\
			& & & \textbf{0.3\% (100)} & \textbf{1\% (500)} & \textbf{14\% (5K)} & \textbf{50\% (17K)} & \textbf{all (35K)}\\
			\hline
			\multirow{2}{*}{MV} 
			& MV-CSS \cite{Sermanet18} & \color{red}{\xmark} &  216.15 & 206.08 & 203.24 & 202.65 & 202.52 \\
			& NSD \cite{Rhodin19a} & \color{green}{\cmark}& \textbf{140.92} &  \textbf{115.37} &  \textbf{94.48} &  \textbf{91.01} &  \textbf{88.89} \\
			\hline
			\multirow{3}{*}{SV} & CSS \cite{Anand19, Chen20a} & \color{red}{\xmark} & 230.77 & 215.96 & 197.22 & 195.70 & 194.69 \\
			& DrNet \cite{Denton17} & \color{green}{\cmark} & 158.81 & 126.18 & 108.21 & 106.41 & 100.38 \\
			& Ours & \color{green}{\cmark} & \textbf{149.32} &  \textbf{122.44} &  \textbf{100.23} &  \textbf{97.38} & \textbf{95.39} \\
			\hline
		\end{tabular}
	}
	\caption{\small {\bf Comparison on the PoseTrack H36M test-set}. The top rows characterize the unsupervised multi-view (MV) models and the bottom rows the single-view (SV) ones. The last 5 columns show N-MPJPE results when using different percentage of labeled 3D pose samples. The value in parenthesis are the corresponding number of images with labels.
	}
	\label{tab:h36}
\end{table}

As discussed in Section~\ref{sec:related}, some 2D keypoint estimation approaches~\cite{Zhang18b,thewlis19,lorenz19, jakab20} separate features for pose and appearance, as we do. For comparison purposes, we therefore also evaluate our extracted time-variant features for 2D keypoint estimation, using the same setup as in Table 1 of~\cite{jakab20}, that is,  we train a single linear layer from the time-variant features to the 2D keypoints. As in~\cite{jakab20}, we use subjects 1,5,6,7,8 for training and 9,11 for testing. As can be seen in Table \ref{tab:h36_kpts}, our model outperforms the other approaches, while it does not use any ground-truth bounding box locations for object detection.


\begin{table}[h]
   \resizebox{.33\linewidth}{!}{
	\begin{minipage}[t]{.2\textwidth}\vspace*{0pt}%
	    \vspace{.5cm}
		\begin{tabu} to .8 \linewidth{| l | c |} \hline
			\textbf{Model} & \textbf{\%-MSE Error} \\ \hline
			Thewlis et al.~\cite{thewlis19} & 7.51 \\
			Zhang et al.~\cite{Zhang18b} & 4.14 \\
			Lorenz et al. ~\cite{lorenz19} & 2.79 \\ 
			Jakab et al.~\cite{jakab20} & 2.73 \\
			Ours & \textbf{2.53} \\
			\hline
		\end{tabu}
	\end{minipage}}  \hfill
	\begin{minipage}[t]{.27\textwidth}
	\setlength{\abovecaptionskip}{0pt}%
	\caption{\small {\bf Comparison against state-of-the-art 2D keypoint estimation models on H36M dataset.} \%-MSE normalized by image size is reported. All models predict 32 keypoints on 6 actions of wait, pose, greet, direct, discuss, and walk.}%
	\label{tab:h36_kpts}
\end{minipage}
	
\end{table}

\subsubsection{MPI Dataset.}

In this dataset as the S1 to S8 videos are multi-view and the TS1 to TS4 ones are single-view, we can only evaluate NSD in the MV-data setting. We report results in Table \ref{tab:mpi}, where we compare with the best-performing models of Table~\ref{tab:h36}. Our model again performs closely to NSD in this setting with the advantage that, because our approach is single-view, we can also train on the TS1 to TS4 videos, thus further improving performance. In a practical setting, this means that we could exploit additional videos as they become available, whether or not they are multi-view, something NSD cannot do.


\begin{table}[ht]
	\centering
	\resizebox{1\linewidth}{!}{
		\begin{tabular}{|c|l|ccccc|}
			\hline
			& \multirow{2}{*}{\textbf{Model}}&  \multicolumn{5}{c|}{\textbf{Percentage of Labeled Pose Data}} \\
			& &  \textbf{.02\% (171)} & \textbf{.1\% (856)} & \textbf{.2\% (1712)} & \textbf{5\% (4K)} & \textbf{2\%(17K)}\\
			\hline
			\multirow{2}{*}{MV-data} 
			& NSD \cite{Rhodin19a}   & \textbf{256.34} & \textbf{241.12} & \textbf{239.81} & \textbf{237.56} & \textbf{235.61} \\
			& Ours  & 261.99 & 245.09 & 241.34 & 239.48 & 236.85 \\
			\hline
			\multirow{2}{*}{All-data} 
			& DrNet \cite{Denton17} & 253.47 & 243.87 & 233.09 & 228.09 & 218.18 \\
			& Ours & \textbf{235.62} & \textbf{219.72} & \textbf{211.26} &  \textbf{202.60} & \textbf{192.11} \\
			\hline
		\end{tabular}
	}
	\caption{\small {\bf Comparison on MPI-INF-3DHP.} N-MPJPE results on pose prediction as a function of the quantity of annotated data. The top rows show results in the MV-data setting, where the self-supervised models have access only to multi-view data. The bottom rows show results in the All-data setting where self-supervised models have access to both multi-view and single-view data. Note that, in this scenario, our results are now better than those of NSD because it can also exploit single-view data.}
	\label{tab:mpi}
	\vspace{-0.2cm}
\end{table}

\subsubsection{Diving Dataset}

We compare with NSD in the MV-data setting and with DrNet in the All-data setting. As can be seen in Table~\ref{tab:diving}, our single-view approach outperforms DrNet, performs on par with multi-view NSD in the MV-data setting and better in the All-data setting. In other words, because our approach can benefit from both multi- and single-view data, it can leverage more of the available images. 

As discussed above, on this dataset, we use the tracking loss presented in Section \ref{sec:gravity} for pre-training purposes for all models, including the baselines. Otherwise, it would take longer for them to detect the subject. The tracking loss reduces the required number of training iterations from 50K to 5K, and thus the training time, saving 38 training hours on a Tesla V100 GPU. Table~\ref{tab:diving} features results for an additional variant of our model, Ours (no $\mathcal{L}_{track}$), which is trained from scratch without gravity pre-training using the same total number iterations as our full method with pre-training. The pose estimation error increases by 21\%. 



\begin{table}[ht]
	\centering
	\resizebox{1\linewidth}{!}{
		\begin{tabular}{|c|l|ccccc|}
			\hline
			\multirow{2}{*}{\textbf{Model}} &  & \multicolumn{5}{c|}{\textbf{Percentage of Labeled Pose Data}} \\
			&&  \textbf{5\% (120)} & \textbf{10\% (241)} & \textbf{20\% (483)} & \textbf{50\% (1209)} & \textbf{100\%(2419)}\\
			\hline
			\multirow{2}{*}{MV-data} & NSD \cite{Rhodin19a}   & 270.42 & 238.23 & 221.25 & 200.28 & 195.13 \\
			& Ours & 279.82 & 242.78 & 224.45 & 205.89 & 197.40 \\ \hline
			\multirow{3}{*}{All-data} 
			& DrNet \cite{Denton17} & 320.07 & 290.93 & 271.87 & 258.14 & 248.31 \\
			& Ours (no $\mathcal{L}_{track}$) & 296.22 & 268.88 & 255.02 & 236.49 & 225.77 \\
			& Ours  & \textbf{265.09} & \textbf{232.66} & \textbf{216.14} & \textbf{197.82} & \textbf{186.45} \\
			\hline
		\end{tabular}
	}
	\caption{\small {\bf Comparison on the Diving dataset (Diver-split).} N-MPJPE results as a function of the quantity of annotated data used to train the pose predictor. Top two rows show models trained only on multi-view data. The bottom rows show models trained using both single-view and multi-view data.}
	\label{tab:diving}
\end{table}

\subsubsection{Ski Dataset}

\begin{table*}[ht]
	\centering
	  \resizebox{.8\linewidth}{!}{
		\begin{tabular}{|l|c|c|c|c|ccccc|}
			\hline
				\multirow{2}{*}{\textbf{Model}} &
			\multirow{2}{*}{\textbf{Decodes?}} & \multirow{2}{*}{\textbf{uses STN?}} & \multirow{2}{*}{\textbf{contrastive loss}} & \multirow{2}{*}{\textbf{latent split?}} &  \multicolumn{5}{c|}{\textbf{Percentage of Labeled Pose Data}} \\
			& & &  & & \textbf{0.3\% (100)} & \textbf{1\% (500)} & \textbf{14\% (5K)} & \textbf{50\% (17K)} & \textbf{all (35K)}\\
			\hline
			DSL & \color{red}{\xmark} & \color{red}{\xmark} &  DSL & \color{red}{\xmark} & 227.24 &  220.64 & 198.34 & 193.08 & 191.21 \\
			
			DSL-STN& \color{red}{\xmark} & \color{green}{\cmark} &  DSL & \color{red}{\xmark} & 236.32 & 211.85 & 202.57 & 199.16 & 198.22 \\
			
			 DSL-STN-Dec & \color{green}{\cmark} & \color{green}{\cmark} &  DSL & \color{red}{\xmark} &  185.75 & 159.34 & 135.93 & 129.62 & 127.52 \\
			
			AE-STN-Dec & \color{green}{\cmark} & \color{green}{\cmark} & None & \color{red}{\xmark}  & 187. 25 & 161.53 & 130.21 & 117.044 & 114.73 \\

			CSS-STN-Dec-Split & \color{green}{\cmark} & \color{green}{\cmark} &  CSS & \color{green}{\cmark} & 163.61 & 137.52&  110.61 & 104.23 & 99.32 \\
			
			DSL-STN-Dec-Split (Ours) & \color{green}{\cmark} & \color{green}{\cmark} &  DSL & \color{green}{\cmark} &  \textbf{149.32} &  \textbf{122.44} &  \textbf{100.23} &  \textbf{97.38} & \textbf{95.39} \\
			\hline
		\end{tabular}
	}
	\caption{\textbf{Ablation Study.} The last row shows our full model with all features applied. In the first two rows we apply the DSL contrastive loss without decoding and latent split. In the third row, we keep all the features except we do not split the latent components. In the fourth row, we apply only detection and reconstruction without any application of contrastive loss. In the fifth row, we only change the contrastive loss of our full model from DSL to CSS. In all settings, we observe a performance drop, however, the models without decoding suffer the most.}
	\label{tab:ablation}
\end{table*}

To check the performance of our model given a changing background, we evaluate our model on a ski dataset~\cite{Ostrek19} where the camera follows skiers down the slope. 
On this dataset, we obtain an N-MPJPE error of 113.20 compared to 180.31 for the single-view approach of DrNet.

In Fig.~\ref{fig:ski_big} we present qualitative results, with input images, the corresponding estimated backgrounds, and the background-subtracted images. Note that, because of noise, background subtraction does not yield images containing only the subject of interest. In rows 4 to 6 we show different thresholdings of the background-subtracted inputs. Note that thresholding by itself cannot eliminate all the background noise and the model needs to learn to focus on the subject and ignore the remaining background. We further show samples of detected bounding boxes, reconstructed images, decoded masks, and 3D pose estimates. 

\subsection{Analysis} 
\label{sec:analysis}

\subsubsection{Ablation Study} 

\begin{figure}[t]
	\centering
	\includegraphics[width=.4\textwidth]{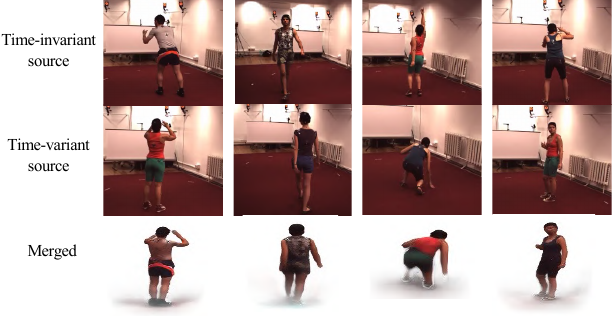}
	\caption{\small {\bf Swapping time-invariant and time-variant components}. In each column, the time-invariant component is taken from the image on the top, and the time-variant component is taken from the image on the bottom and concatenated to generate the image shown in bottom, using the decoder.}
	\label{fig:swap}
\end{figure}

To analyze the different components of our model, we perform an ablation study on H36M by removing different model components and report our results in Table~\ref{tab:ablation}.

In the first two rows (DSL, DSL-STN), we only apply the DSL contrastive loss, without latent splitting and reconstruction. Compared to our full approach in the last row of the table, the accuracy is degraded by a factor of almost 2. Since these two models do not reconstruct the image, they cannot learn the time-invariant components and all the latent components are time-variant. This being the case, the basic contrastive approach does not suffice. 

In the third row, we add reconstruction of the input to the DSL-STN model but without splitting the latent variables. The error of DSL-STN-Dec in the last column improves by 36\% compared to DSL-STN. Reconstructing the image helps the model to both detect the subject of interest and to capture more details. 

In the final row, we report again the results of the full model, which adds latent splitting to the DSL-STN-Dec model of the third row. It improves pose estimation by 25\% compared to DSL-STN-Dec, showing splitting the latent space is beneficial in learning better time-variant features, as the model can apply the contrastive loss only to this component. Compared to the basic DSL approach, we improve by 50\%. In other words, detection, reconstruction and latent vector splitting all help deliver the best possible accuracy.

We also compare against AE-STN-Dec, which only applies detection and decoding, without any contrastive loss or latent splitting. The error in the last column increases by about 20\% compared to our full approach. 
Finally, in the fifth row, we change only the DSL loss of Eq.~(\ref{eq:DSL}) to the CSS loss of Eq.~(\ref{eq:cssLoss}) and still observe an increase in error by 4\% in the last column. This gap is even more noticable in previous columns of the table, that is, when using fewer annotations. We observe an error increase of about 10\% for the columns corresponding to 0.3\%, 1\%, and 14\% of the data, and of about 7\% for the column corresponding to 50\%.

In short, all of the proposed modules have a role and disabling any of them reduces performance.

\subsubsection{Validity of our Distance-Based Similarity Loss} 

\begin{figure}[h!]
	\centering
	\begin{subfigure}[b]{0.25\textwidth}
		\centering
		\includegraphics[width=\textwidth]{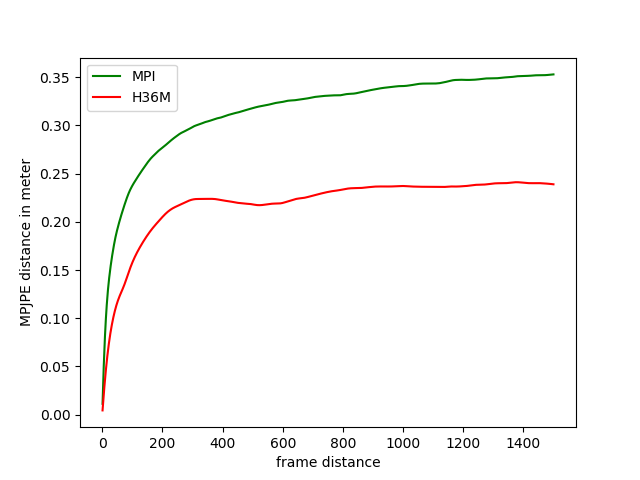}
	\end{subfigure}
	\hspace{-0.5cm}
	\begin{subfigure}[b]{0.25\textwidth}
		\includegraphics[width=\textwidth]{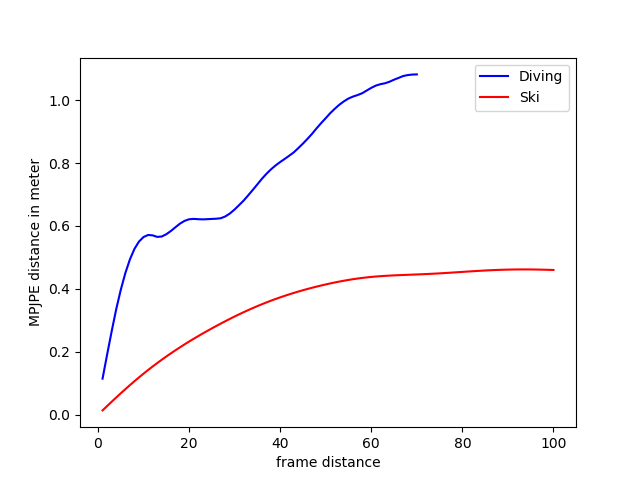}
	\end{subfigure}
	\caption{\small Average MPJPE distance of pelvis-centered poses at different frame distances.}
	\label{fig:mpjpe-dist}
\end{figure}

To verify whether the DSL loss proposed in Section \ref{sec:CSS} makes sense, we measure how much 3D pose differs over time in videos. Fig.~\ref{fig:mpjpe-dist} shows the average MPJPE distance over the training set of pelvis-centered poses as a function of frames temporal distances. For different datasets, the pose distance is different  because the actions are different. For example, in the Diving dataset, pose distances change more quickly and reach a higher level than in the other datasets, mostly because the divers rotate during their dives. On average, while poses are similar in nearby frames they get more dissimilar as the temporal distance increases before plateauing. This trend matches the assumption of how frames change over time that underpins our formulation of the  DSL loss of Eq.~(\ref{eq:DSL}). 

\subsubsection{Disentanglement}
\label{sec:exp_disentangle}

For us disentanglement is not a goal by itself but rather a means to an end. However, to verify the disentanglement of the time-variant (TV) and time-invariant (TI) components, we applied KMeans clustering separately on the TV and TI features of our model on the H36M test-set and compared them to the subject classes. We obtained 52.3\% and 95.9\% classification accuracy for TV and TI, showing that TI captures identity while TV does not. We also trained a pose model on the TI features of H36M and obtained an N-MPJPE of 201.3 for the 100\%-data case, compared to 95.39 for TV (presented in Table \ref{tab:h36}), showing that TV is almost twice better than TI for pose estimation. Hence, disentanglement is indeed happening.



\begin{table}[ht]
	\centering
	\resizebox{1\linewidth}{!}{
		\begin{tabular}{|l|l|c|c|c|c|}
			\hline
			\textbf{1st-phase} & \textbf{2nd-phase} & \textbf{Train-set} & \textbf{MPJPE} & \textbf{N-MPJPE} & \textbf{P-MPJPE} \\
			\hline
			Resynthesis (w STN) & $\bI_{\TV}$ & H36M & 10.17 & 10.09 & 7.38 \\
			Resynthesis (w/o STN) & $\bI_{\TV}$ & H36M & 10.03 & 9.93 & 7.49 \\
			\hline
			Resynthesis (w/o STN) + 2D & $\bI_{\TV}$ & H36M & 7.45 & 7.41 & 5.73 \\
			Resynthesis (w/o STN) + 2D & $\bI_{\TV}$ + 2D & H36M & 7.33 & 7.29 & 5.61 \\
			\hline
			Resynthesis (w/o STN) + 2D \textbf{*} & $\bI_{\TV}$ & MPI & 7.46 & 7.42 & 5.72 \\
			Resynthesis (w/o STN) + 2D \textbf{*} & $\bI_{\TV}$ + 2D & MPI & 7.39 & 7.35 & 5.68 \\
			\hline
			Resynthesis (w/o STN) + 2D \textbf{*} & 2D & H36M & 6.42 & 6.37 & 5.01	 \\
			Resynthesis (w/o STN) + 2D \textbf{*} & $\bI_{\TV}$ & H36M & 6.26 & 6.23 & 4.82 \\
			Resynthesis (w/o STN) + 2D \textbf{*} & $\bI_{\TV}$ + 2D & H36M & 6.01 & 5.97 & 4.60 \\
			\hline
			Resynthesis (w/o STN) + 2D \textbf{*} & $\bI_{\TV}$ + 2D alpha & H36M & \textbf{5.69} & \textbf{5.65} & \textbf{4.36} \\
			\hline
		\end{tabular}
	}
	\caption{\small {\bf Comparison on H36M test-set (in cm)}. 
	First two rows apply image re-synthesis with and without detection in the first phase of training, where in the latter the cropped image is passed to the model. In the 3rd and 4th rows 2D is used together with image re-synthesis in the first phase of training. The bottom rows with $*$ show models where the ResNet50 backbone is replaced with a Resnet 101. The 5th and 6th rows show results where the model is trained on MPI in the first phase and then tested on H36M in the 2nd phase. When 2D is used in the 2nd phase, it is the prediction of the model itself. In the last row, however, we pass $\bI_{\TV}$ together with 2D pose predicted by Alphapose to 3D pose estimation model in the 2nd phase. All models use a ResNet MLP in the 2nd phase.
	}
	\label{tab:h36_weak_ablation}
\end{table}


\begin{table}[ht]
	\centering
	\resizebox{1\linewidth}{!}{
		\begin{tabular}{|l|l|c|c|c|c|}
			\hline
			\textbf{1st-phase} & \textbf{2nd-phase} & \textbf{Train-set} & \textbf{MPJPE} & \textbf{N-MPJPE} & \textbf{P-MPJPE} \\
			\hline
			Resynthesis & $\bI_{\TV}$ (2-hid MLP) & MPI & 18.06 & 17.49 & 12.31  \\
			Resynthesis & $\bI_{\TV}$ (ReNet MLP) & MPI & 16.80& 16.37 & 11.15 \\
			\hline
			Resynthesis + 2D \textbf{*} & $\bI_{\TV}$  (ReNet MLP) & H36M & 11.50 & 11.40 & 7.54 \\
			Resynthesis + 2D  \textbf{*} & $\bI_{\TV}$ + 2D (ReNet MLP) & H36M & 11.39 & 11.29 & 7.49\\
			\hline
			Resynthesis + 2D \textbf{*} & $\bI_{\TV}$ (ReNet MLP) & MPI & 10.21 & 10.15 & 6.54  \\
			Resynthesis + 2D \textbf{*} & $\bI_{\TV}$ + 2D (ReNet MLP) & MPI & \textbf{9.30} & \textbf{9.24} & \textbf{5.95} \\
			\hline
		\end{tabular}
	}
	\caption{\small {\bf Comparison on MPI-INF-3DHP test-set (in cm)}. 
	First two rows apply only image re-synthesis, where the first row uses 2-hid MLP for 3D pose regression. The last rows with $*$ show models where the ResNet50 backbone is replaced with a Resnet 101. The 3rd and 4th rows show results where the model is trained on H36M in the first phase and then tested on MPI in the 2nd phase.
    }
	\label{tab:mpi_weak_ablation}
\end{table}


\begin{table}[ht]
	\centering
	\resizebox{1\linewidth}{!}{
		\begin{tabular}{|l|l|c|c|c|c|}
			\hline
			\textbf{dataset} & \textbf{1st-phase} &\textbf{2nd-phase} & \textbf{MPJPE} & \textbf{N-MPJPE} & \textbf{P-MPJPE} \\
			\hline
			Diver-split & Resynthesis & $\bI_{\TV}$  & 16.4 & 15.7 & 10.7  \\
			Diver-split & Resynthesis + 2D & $\bI_{\TV}$  & 15.2 & 14.1 & 9.43 \\
			\hline
			Dive-split & Resynthesis & $\bI_{\TV}$  & 9.19 & 8.75 & 5.81 \\
			Dive-split & Resynthesis + 2D & $\bI_{\TV}$ &  \textbf{8.53} &  \textbf{8.26} &  \textbf{5.60} \\
			\hline
		\end{tabular}
	}
	\caption{\small 
	{\bf Comparison on the Diving dataset (in cm)}. 
	The first two rows compare the results in the diver-split setting, whereas the last two rows evaluate the dive-split setup. All models use a ResNet50 backbone and use a 6-layer MLP as pose model.
    Only the time-variant features are used for pose estimation to compare each pair of rows with a similar input.}
	\label{tab:dive_weak_ablation}
\end{table}


\begin{table}[ht]
	\centering
	\resizebox{.8\linewidth}{!}{
		\begin{tabular}{|l|l|c|c|c|}
			\hline
			&  \textbf{Model}  & \textbf{MPJPE} & \textbf{N-MPJPE} & \textbf{P-MPJPE} \\
			\hline
			\multirow{3}{*}{Unsup} 
			& Rhodin \cite{Rhodin18b} & - & 115.0 & - \\
			& Kundu \textit{et al.} \cite{Kundu20a} & \textbf{99.2} & - & - \\
			&  Ours & 100.3 & \textbf{99.3} & \textbf{74.9} \\
			\hline
			
			\multirow{11}{*}{Weakly-Sup}
	
			& AIGN \cite{Tung17a} & - & - & 79.0 \\
			
			& MCSS \cite{Mitra20} & 94.25 & 92.60 & 72.48 \\ 
			
			& Chen et al. \cite{Chen19g} & - & - & 68.0 \\
			
			& HMR \cite{Kanazawa19a} & - &  - & 66.5 \\
			
			& RepNet \cite{Wandt19} & 89.9 & - & 65.1 \\
			
			& Wang \textit{et al.} \cite{Wang19k} &  86.4 & - &  62.8 \\
						
			& Kolotouros \textit{et al.} \cite{Kolotouros19} & - & - & 62.0 \\
			
			& Iqbal \textit{et al.} \cite{Iqbal20} & 67.4 & 64.5  & 54.5 \\ 
			
			& CanonPose  \cite{Wandt21} & 74.3 & - & 53.0 \\ 
			
			& Kundu \textit{et al.} \cite{Kundu20a} & 62.4 & - & - \\

			& Ours & \textbf{60.1} & \textbf{59.7} & \textbf{46.0} \\
			\hline

			
		\end{tabular}
	}
	\caption{\small {\bf Comparison with models on H36M test-set (in mm)}. 
	Unsupervised methods do not use any sort of labels in the feature extraction phase. Weakly-supervised methods use either 2D labels or a pre-trained 2D model.
	}
	\label{tab:h36_weak}
\end{table}


\begin{table}[ht]
	\centering
	\resizebox{0.8\linewidth}{!}{
		\begin{tabular}{|l|l|c|c|c|}
			\hline
			&  \textbf{Model}  & \textbf{MPJPE} & \textbf{N-MPJPE} & \textbf{P-MPJPE} \\
			\hline
			
			\multirow{8}{*}{Weakly-Sup}
		
			& Kolotouros \textit{et al.} \cite{Kolotouros19} & 124.8 & - & 80.4 \\
						
			& HMR \cite{Kanazawa19a} & 169.5 &  - & 113.2 \\
							
			& Iqbal \textit{et al.} \cite{Iqbal20} & 113.8 & 102.2  & - \\ 
			
			& CanonPose  \cite{Wandt21} & 104.0 & - & 70.3 \\ 
			
			& Chen et al. \cite{Chen19g} & - & - & 71.1 \\
				
			& RepNet \cite{Wandt19} & 97.8 & - &  \\			
			
			& Kundu \textit{et al.} \cite{Kundu20a} & 93.9 & - & - \\

			& Ours & \textbf{93.0}
 & \textbf{92.4} & \textbf{59.5}\\
			\hline
			
			
		\end{tabular}
	}
	\caption{\small {\bf Comparison with models on MPI-INF-3DHP test-set (in mm)}. 	
	Weakly-supervised methods use either 2D labels or a pre-trained 2D model.
	}
	\label{tab:mpi_weak}
\end{table}


\begin{table}[ht]
	\centering
	\resizebox{1\linewidth}{!}{
		\begin{tabular}{|l|l|c|c|c|c|}
			\hline
			\textbf{Model} & \textbf{2nd-phase} & \textbf{MPJPE} & \textbf{N-MPJPE} & \textbf{P-MPJPE} \\
			\hline
			ImageNet + 2D &  $\bI_{\TV}$  + 2D & 18.38 & 18.07 & 11.44  \\
			Resynthesis + 2D (Ours) & $\bI_{\TV}$ + 2D  & 13.35 & 13.29 & 9.20 \\
			\hline
			ImageNet + 2D \textbf{*} &  $\bI_{\TV}$ + 2D  & 16.47 & 16.29 & 11.02  \\
			Resynthesis + 2D \textbf{*} (Ours) & $\bI_{\TV}$ + 2D & \textbf{10.42} & \textbf{10.38} & \textbf{7.47} \\
			\hline
		\end{tabular}
	}
	\caption{\small 
	{\bf Results on 3DPW (in cm)}. 
	We compare the impact of our pre-training approach to the ImageNet pre-trained models on in-the-wild images. The first two rows use ResNet50 as backbone, whereas the bottom two rows, indicated by *, use ResNet101. 
	}
	\label{tab:3dpw}
	\vspace{-0.3cm}
\end{table}


\begin{figure}[h!]
	\centering
	\begin{subfigure}[b]{0.105\textwidth}
	\includegraphics[width=\textwidth]{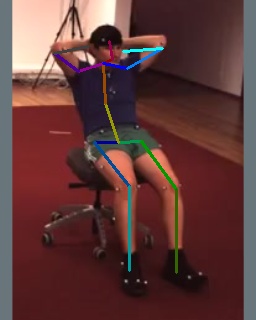} 
	\end{subfigure}
	\hspace{0.01em}
	\begin{subfigure}[b]{0.105\textwidth}
		\includegraphics[width=\textwidth]{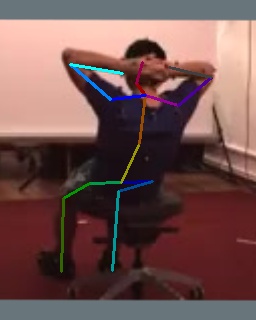} 
	\end{subfigure}
	\hspace{0.01em}
	\begin{subfigure}[b]{0.105\textwidth}
	\includegraphics[width=\textwidth]{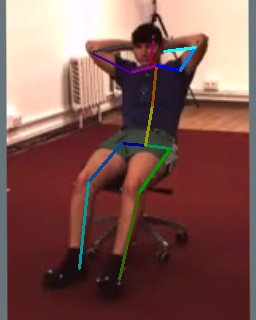}
	\end{subfigure}
	\vspace{0.2cm}

	\begin{subfigure}[b]{0.105\textwidth}
		\includegraphics[width=\textwidth]{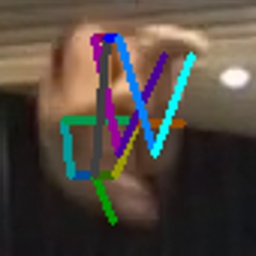}  \caption*{Original View}
	\end{subfigure}
	\hspace{0.01em}
	\begin{subfigure}[b]{0.105\textwidth}
		\includegraphics[width=\textwidth]{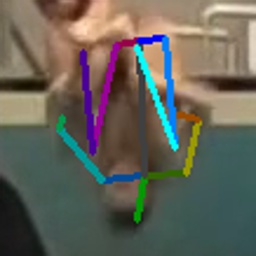} \caption*{Side-View 1}
	\end{subfigure}
	\hspace{0.01em}
	\begin{subfigure}[b]{0.105\textwidth}
		\includegraphics[width=\textwidth]{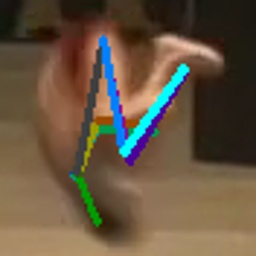} \caption*{Side-View 2}
	\end{subfigure}
	\caption{\small Side-view visualization. For each triplet, the first image shows the prediction of the model on the input view, while side-views 1 and 2 show projection of the 3D pose to other views.}
	\label{fig:side_view_main}
	\vspace{-0.4cm}
\end{figure}

In Fig.~\ref{fig:swap}, we show results obtained by merging time-variant and time-invariant components from different sources to reconstruct new images. We observe that the model correctly splits information into time-variant and time-invariant components, which it can then use to synthesize a new image.

\subsection{Leveraging Weak Labels} 
\label{sec:exp_weak_labels}

We evaluate how leveraging 2D poses, when available as described in Section \ref{sec:weak_labels}, can help improve the learned $\bI_{\TV}$ features.  Tables \ref{tab:h36_weak_ablation} shows the impact of using 2D poses in the feature extraction phase on H36M dataset. In particular, comparing 2nd and 3rd rows of the table show using 2D pose can improve MPJPE by about 2.5 cm. In the 2nd row we provide cropped images to the network. Comparing the first two rows show the models perform almost similarly with and without detection, indicating the efficiency of our unsupervised detection in the first row. 

In the bottom rows, we replace the ResNet50 backbone with a ResNet101, which further improves the accuracy. This indicates a more sophisticated architecture can yield improved performance. In the final row we take 2D poses from pre-trained Alphapose model \cite{Fang17a}, which is fine-tuned on H36M, and pass it together with $\bI_{\TV}$ to 3D pose estimation network.  Rows 7th to 9th also show that adding 2D in the 2nd phase to $\bI_{\TV}$ yields the best results rather than using them individually. 


\begin{figure*}[h!]
	\centering
	\begin{subfigure}[b]{0.25\textwidth}
		\centering
		\includegraphics[width=\textwidth]{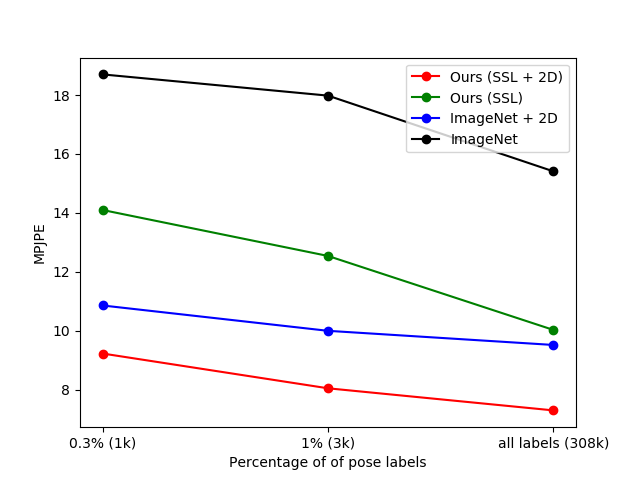}
	\end{subfigure}
	\hspace{-0.5cm}
	\begin{subfigure}[b]{0.25\textwidth}
		\includegraphics[width=\textwidth]{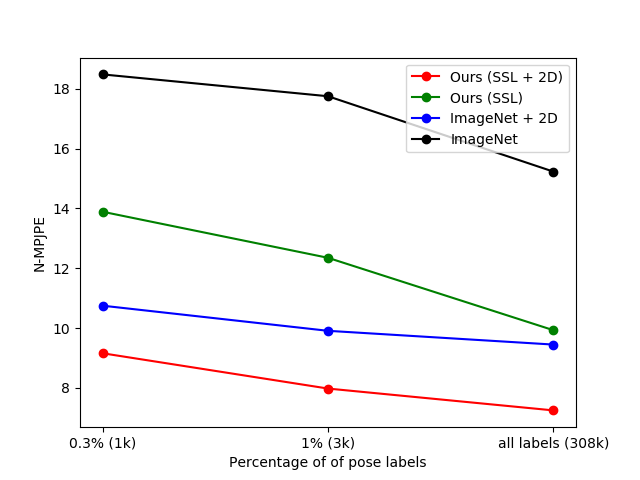}
	\end{subfigure}
	\begin{subfigure}[b]{0.25\textwidth}
	\includegraphics[width=\textwidth]{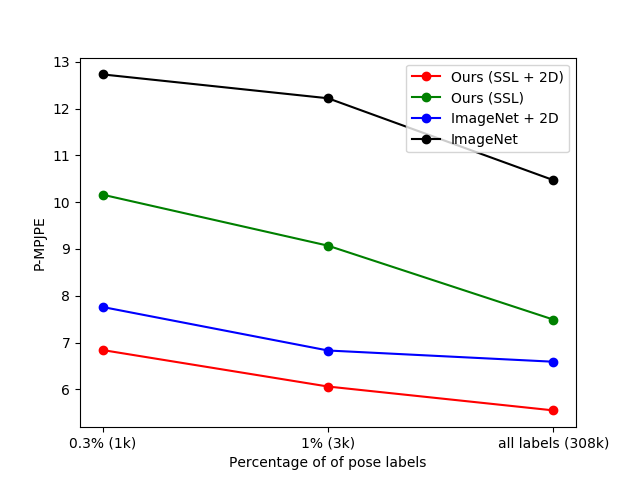}
	\end{subfigure}
	\caption{\small 
		Comparison of extracted feature by Ours versus supervised ImageNet on H36M test-set using a ResNet50 architecture. The three plots from left to right show results on MPJPE, NMPJPE, and PMPJPE (in cm) for different percentage of labeled 3D data. Ours SSL and SSL+2D are variants trained without and with 2D labels. For ImageNet variants we take the features of the layer before the classification one. In ImageNet+2D we also fine-tune this network on 2D labels for a fair comparison with Ours (SSL+2D). 
	}
	\label{fig:imgNet_tv_comp}
	\vspace{-0.5cm}
\end{figure*}

\begin{figure*}[h!]
	\centering
	\begin{subfigure}[b]{0.25\textwidth}
		\centering
		\includegraphics[width=\textwidth]{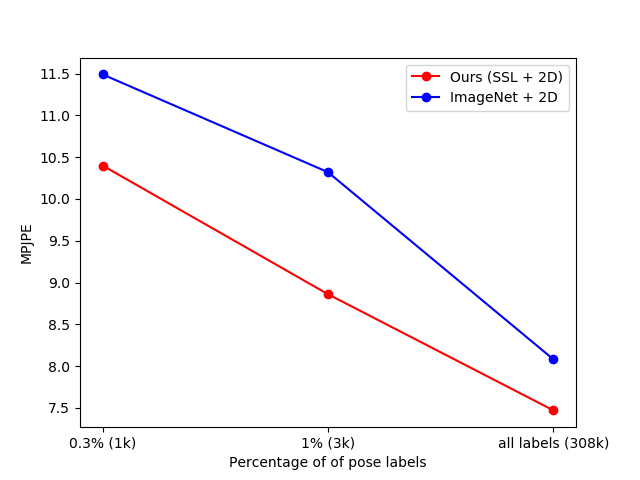}
	\end{subfigure}
	\hspace{-0.5cm}
	\begin{subfigure}[b]{0.25\textwidth}
		\includegraphics[width=\textwidth]{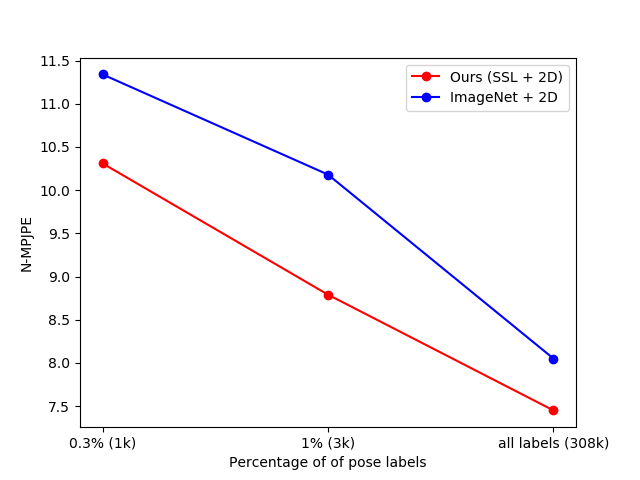}
	\end{subfigure}
	\begin{subfigure}[b]{0.25\textwidth}
	\includegraphics[width=\textwidth]{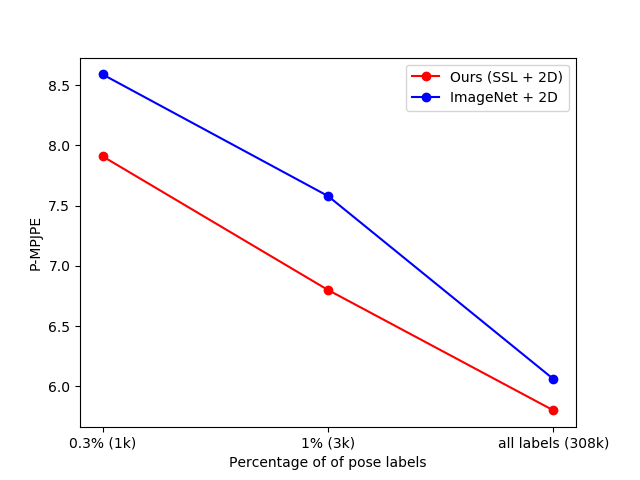}
	\end{subfigure}
	\caption{\small 
		Comparison of the quality of the learned parameters for model (ResNet50 architecture) initialization by Ours versus the supervised ImageNet on H36M test-set. The three plots show results on MPJPE, NMPJPE, and PMPJPE (in cm) for different percentage of 3D labels.
	}
	\label{fig:imgNet_finetune_comp}
	\vspace{-0.5cm}
\end{figure*}

Table \ref{tab:mpi_weak_ablation} similarly shows the impact of 2D on MPI dataset. In the first two rows, where the model only applies re-synthesis, we compare the impact of pose-regressor architecture. The 2nd row shows that using a better ResNet MLP architecture, take from \cite{Martinez17a}, can further boost the performance. The 3rd and 4th rows of Table \ref{tab:mpi_weak_ablation}  as well as 5th and 6th rows of Table \ref{tab:h36_weak_ablation}, show the impact of training on one dataset and testing on another. As expected, the results are about 1 to 2 cm worse than when training and testing on the same dataset. Nevertheless, the results are still decent on the target dataset, which indicates our color augmentation helps the model not overfit on the background or the person.

Table \ref{tab:dive_weak_ablation} shows our results on the diving dataset in the diver-split and dive-split cases. Similarly to the previous datasets, in each setup we obtain better features when using 2D labels. The gap between the two different splits indicates that the model suffers from diver diversity in the diver-split case. Hence, diversity in identities is essential to help the model generalize better. 

\textbf{Comparison with the state-of-the-art models.} 
In Table \ref{tab:h36_weak}  we report results on the test set of H36M dataset and compare against unsupervised models (without any sort of supervision in the feature extraction phase) and also weakly supervised methods that leverage either 2D labels or pre-trained 2D networks. As can be observed our approach compares favorably with respect to these methods. The model of \cite{Kundu20a} leverages some in-house private dataset in addition to an off-the-shelf body-part segmentation, making it not completely unsupervised. In Table \ref{tab:mpi_weak} we compare methods on MPI-INF-3DHP dataset. Similar to H36M our approach obtains better result in the weakly-supervised setup. 

\textbf{Comparison with supervised pre-training.}
Similar to the recent unsupervised learning methods~\cite{He20c, Bao22, He22}, we compare the effectiveness of the features extraction using a ResNet50 backbone with our approach to those obtained with a supervised ImageNet pre-trained model. The results are provided in Figure~\ref{fig:imgNet_tv_comp}. For a fair comparison between our variant that leverages 2D and the ImageNet pre-trained model, we also fine-tune the latter on 2D labels. Ours without 2D improves upon the ImageNet baseline by at least 4.5 cm, while Ours+2D also performs better than the ImageNet+2D variant by 1 to 2 cm in MPJPE. The results clearly show that the features extracted by Ours outperform the ones extracted by a supervised ImageNet model.

To further evaluate the quality of weight transfer, we take the best models in Figure~\ref{fig:imgNet_tv_comp} and use them to initialize networks for end-to-end training on 3D labels. This experiment compares the two models on the quality of the transferred parameters as model initialization.  The results depicted in Figure~\ref{fig:imgNet_finetune_comp} clearly show the advantage of our pre-training.

\textbf{3DPW dataset.}
To evaluate the impact of our feature extraction network on in-the-wild images, we use the 3DPW dataset. We take our best models pre-trained on the H36M dataset in the weakly supervised setup, and compare the quality of the extracted features to the supervised ImageNet variants. For a fair comparison with our Resynthesis+2D models, we also fine-tune the ImageNet models on the H36M 2D labels; without this (as also shown in Figure~\ref{fig:imgNet_tv_comp}) the gap with our approach was even larger. The result are provided in Table \ref{tab:3dpw}. Comparing every pair of rows clearly shows that the features extracted by our approach are much better suited for pose estimation than those obtained with ImageNet pre-trained models.


\section{Limitations} 
While we do not need the cameras to be static, as shown by our experiments on the Ski dataset, we still need a way to obtain a background image. Furthermore, our current formulation can only handle a single subject of interest. We leave the extension to multiple subjects for future work. One approach would be to add depth estimation to the self-supervised model, as proposed in~\cite{Rhodin19a}, and use the resulting depth to reason about occlusions when merging the decoded crops of different subjects to reconstruct the input image.



\section{Conclusion}

In this work, we have presented an unsupervised feature extraction model for monocular RGB videos. It improves upon the standard CSS approaches by applying contrastive learning only to the time-variant components of the latent representations and enforcing latent similarity given the distance between the frames. When combined with detection and an additional image decoding, this allows our model to extract richer features, well suited for capturing transient dynamics in videos, such as human motion.
We have evidenced the benefits of our approach for pose estimation on five benchmark datasets; our approach was shown to outperform other single-view self-supervised learning strategies and to match the performance of multi-view ones.
Finally, we show we can leverage 2D poses to enhance the quality of extracted features in the time-variant component, making 3D pose estimation more accurate.

\section*{Acknowledgment}
This work was supported in part by Innosuisse, the Swiss Innovation Agency. We also would like to thank SwissTiming for their support and cooperation in this project. 

{\small
	\bibliographystyle{IEEEtran}
	\bibliography{string,vision,learning,graphics,biomed}
}


\vspace{-1.3cm}
\begin{small}
	
	\def \bioSpacing {-1.5cm}
	\begin{IEEEbiography}[{\includegraphics[width=1in,height=1.25in,clip,keepaspectratio]{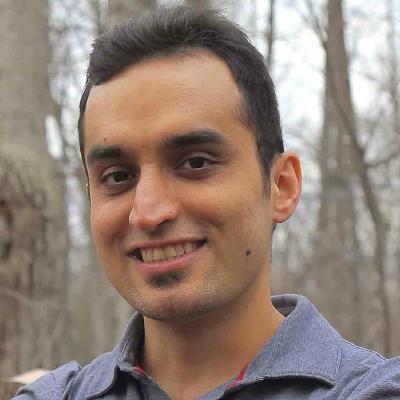}}]{Sina Honari}
	is a post-doctoral reseacher in computer vision laboratory at EPFL, where he works on 3D human pose estimation, multi-view geometry, road-obstacle detection, and animal neural decoding. He received his PhD from University of Montreal and Mila lab, and then was a post-doctoral  researcher at Polytechnique Montreal. His research interests are computer vision and deep learning, in particular self-supervised and semi-supervised methods.
	\end{IEEEbiography}
	\vspace{-1.5cm}

	\begin{IEEEbiography}[\vspace{-1.cm} {\includegraphics[width=1in,height=1.25in,clip,keepaspectratio]{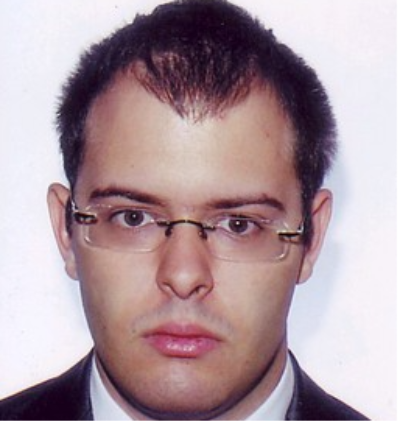}}]
		{Victor Constantin}
		received his M.Sc. degree in Computer Science from EPFL in 2016. He was working as a Research Engineer in Computer Vision Laboratory, EPFL. His research interests include deep learning, 3D pose estimation and garment draping. He is now a machine learning engineer at Uplift Labs.
	\end{IEEEbiography} 	
    \vspace{\bioSpacing}
    \vspace{-0.7cm}

    \begin{IEEEbiography}[{\includegraphics[width=1in,height=1.25in,clip,keepaspectratio]{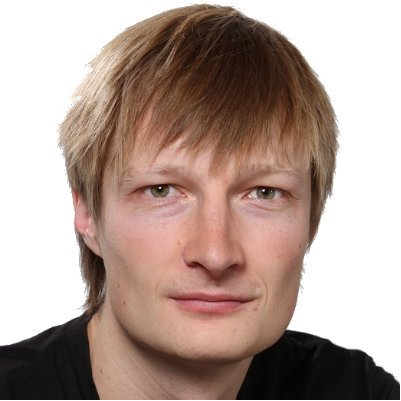}}]{Helge Rhodin}
       is an Assistant Professor at the University of British Columbia. His research interests range from computer graphics and augmented reality, over 3D computer vision, to self-supervised machine learning. Helge received the BSc and MSc degree in Computer science from Saarland University. He graduated with a PhD in 2016 for is work at the Max-Planck Institute for Informatics and was a postdoctoral researcher and lecturer at EPFL.
    \end{IEEEbiography} 
   \vspace{\bioSpacing}
    
	\begin{IEEEbiography}[{\includegraphics[width=1in,height=1.25in,clip,keepaspectratio]{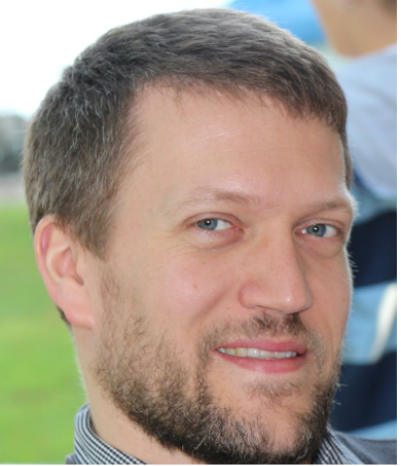}}]{Mathieu Salzmann}
	is a Senior Researcher at EPFL and an Artificial Intelligence Engineer at ClearSpace. Previously, he was a Senior Researcher and Research Leader in NICTA’s computer vision research group, a Research Assistant Professor at TTI-Chicago, and a postdoctoral fellow at ICSI and EECS at UC Berkeley. He obtained his PhD in 2009 from EPFL. His research interests lie at the intersection of machine learning and computer vision.
	\end{IEEEbiography}
	\vspace{\bioSpacing}
	
	\begin{IEEEbiography}[{\includegraphics[width=1in,height=1.25in,clip,keepaspectratio]{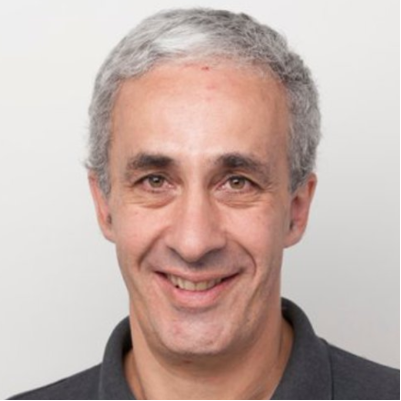}}]{Pascal Fua}
		received an engineering degree from Ecole Polytechnique, Paris, in 1984 and a Ph.D. in Computer Science from the University of Orsay in 1989. He joined EPFL in 1996 as a Professor in the School of Computer and Communication Science. He is the head of the Computer Vision Lab. He has (co)authored over 300 publications in refereed journals and conferences and received several ERC grants. He is an IEEE Fellow and has been an Associate Editor of IEEE Transactions for Pattern Analysis and Machine Intelligence. 
	\end{IEEEbiography}

\end{small}

\newpage
\setcounter{figure}{0}

\setcounter{section}{0}
\renewcommand{\thesection}{S.\arabic{section}}
\renewcommand{\thesubsection}{\thesection.\arabic{subsection}}

\newcommand{\beginsupplementary}{%
	\setcounter{table}{0}
	\renewcommand{\thetable}{S\arabic{table}}%
	\setcounter{figure}{0}
	\setcounter{page}{1}
	\renewcommand{\thefigure}{S\arabic{figure}}%
}

\beginsupplementary
\twocolumn[{%
	\vspace{3. em}
	\centering
	\textbf{\LARGE Temporal Representation Learning on Monocular Videos for 3D Human Pose Estimation  ---	Supplementary Information\\}
	\vspace{3. em}
}
]


\begin{figure*}[thbp]
	\centering
	\includegraphics[width=1\textwidth]{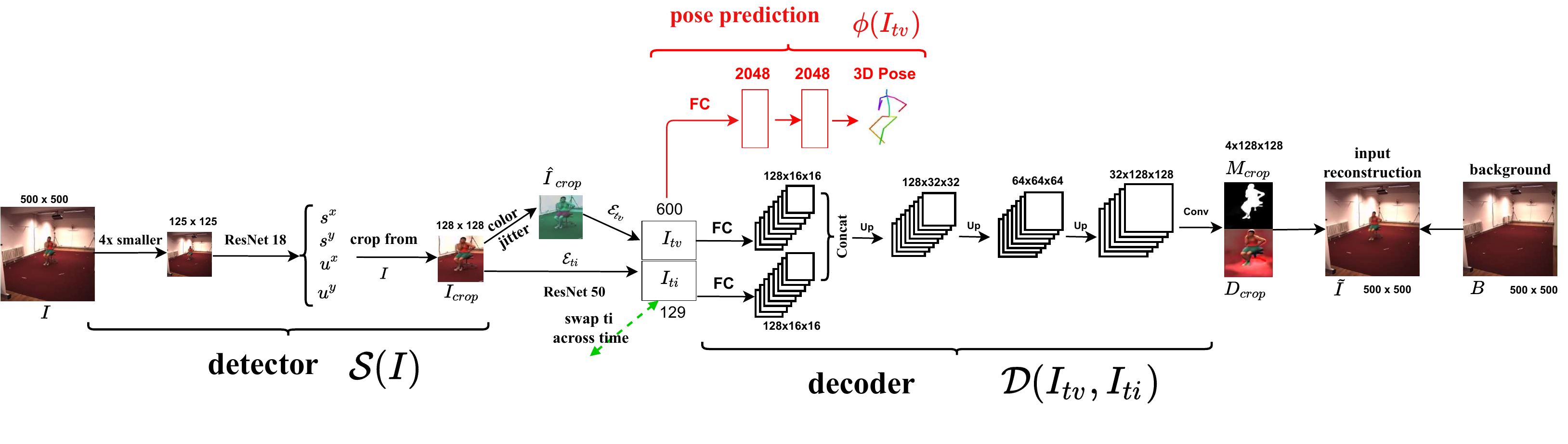}
	\caption{\small \textbf{Model architecture details.} This image depicts the implementation details described in Section \ref{sec:impl}.
	}
	\label{fig:arch_details}
	\vspace{1cm}
\end{figure*}


\begin{figure*}[!htb]
	\centering
	\includegraphics[width=1\textwidth]{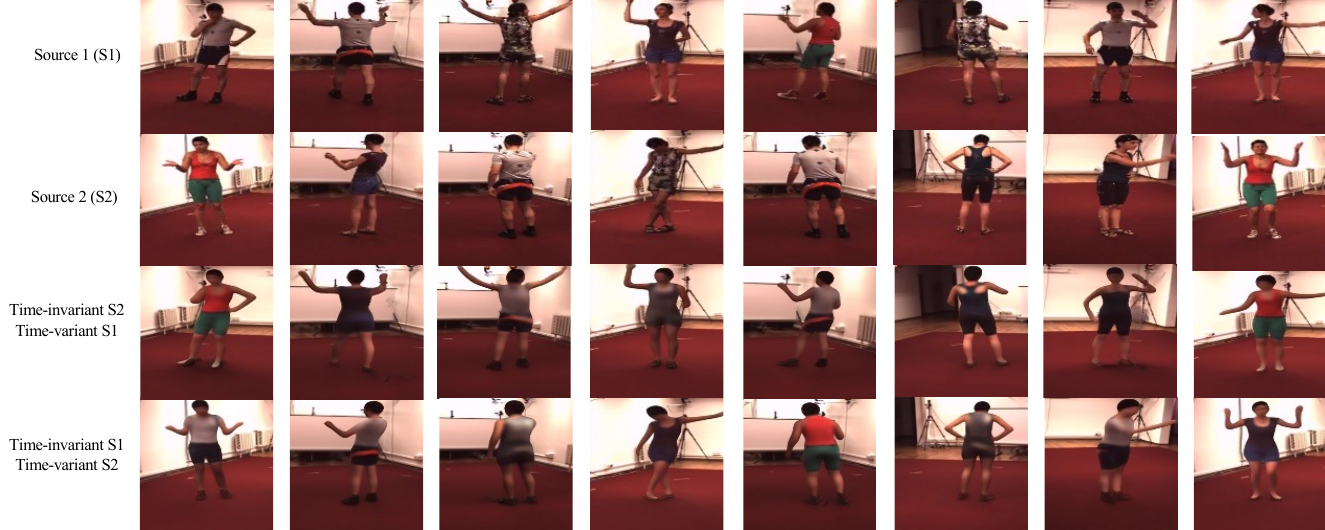}
	\caption{\small {\bf Swapping time-invariant and time-variant components}. Top two rows show two source images S1 and S2. In each column, in the third row the time-invariant component is taken from S2, and the time-variant component is taken from S1 and concatenated to generate the image shown, using the decoder. In the last row, the sources are swapped. As can be observed the model manages to correctly swap time-invariant and time-variant components. However, due to low-dimensional feature representation of the latent components, not all details are maintained. For examples, the clothes details are lost.}
	\label{fig:supp_swap1}
	\vspace{1cm}
\end{figure*}


\begin{figure*}[!htb]
	\centering
	\includegraphics[width=1\textwidth]{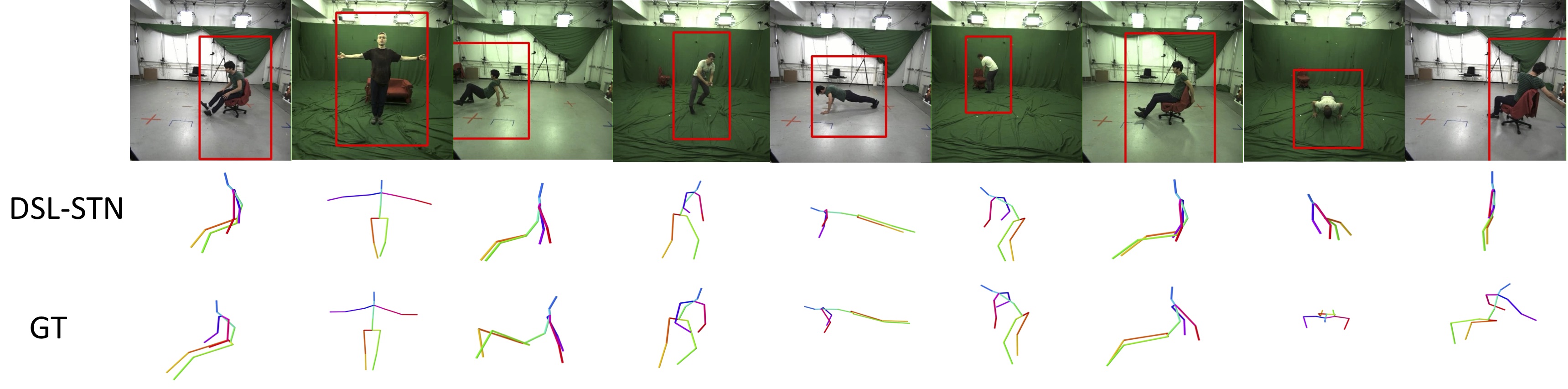}
	\caption{\small {\bf Qualitative results on MPI Dataset}.  Samples are taken from the test set. In the top row, the red  rectangle shows the bounding box detected by our model for image cropping. The two bottom rows, respectively show the pose estimation by our model and the ground truth pose. The last two columns show examples with high error.}
	\label{fig:supp_mpi_pose}
	\vspace{1cm}
\end{figure*}

\begin{figure*}[!htb]
	\centering
	\includegraphics[width=1\textwidth]{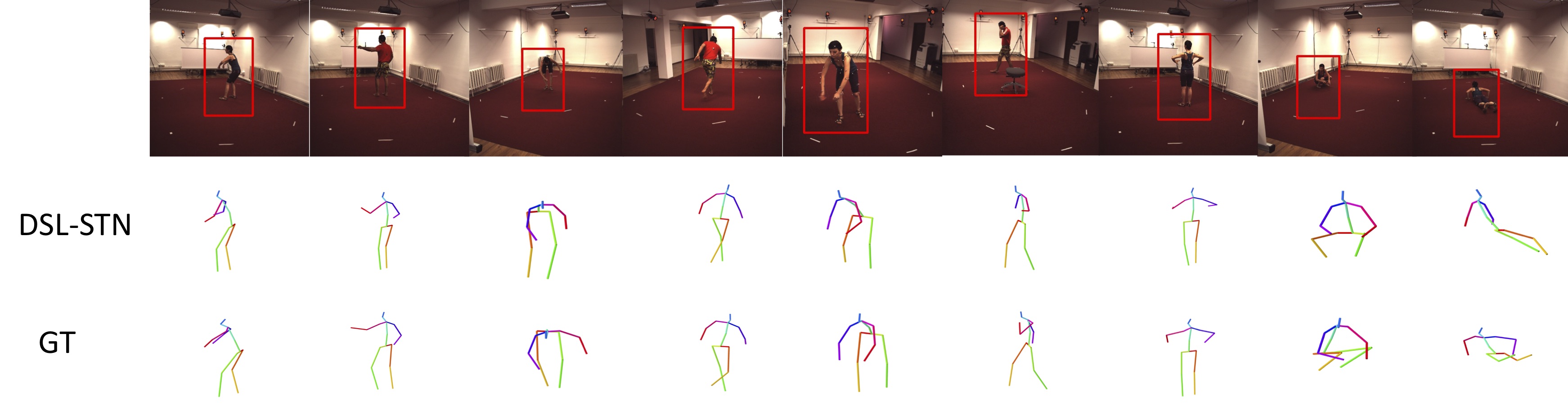}
	\caption{\small {\bf Qualitative results on H36M Dataset}.  Samples are taken from the test set. In the top row, the red  rectangle shows the bounding box detected by our model for image cropping. The two bottom rows, respectively show the pose estimation by our model and the ground truth pose. The last two columns show examples with high error.}
	\label{fig:supp_posetrack}
\end{figure*}

\begin{figure*}[!htb]
	\vspace{1cm}
	\centering
	\includegraphics[width=0.45\textwidth]{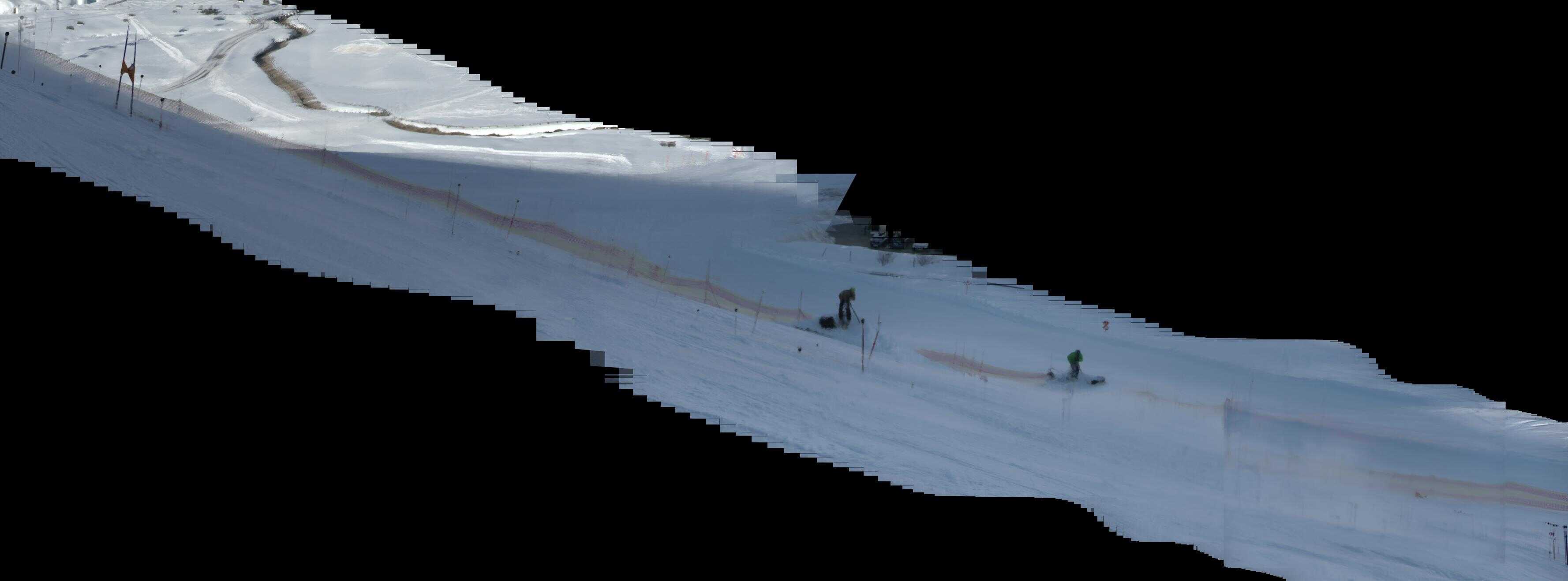}
	\includegraphics[width=0.47\textwidth, height=3.05cm]{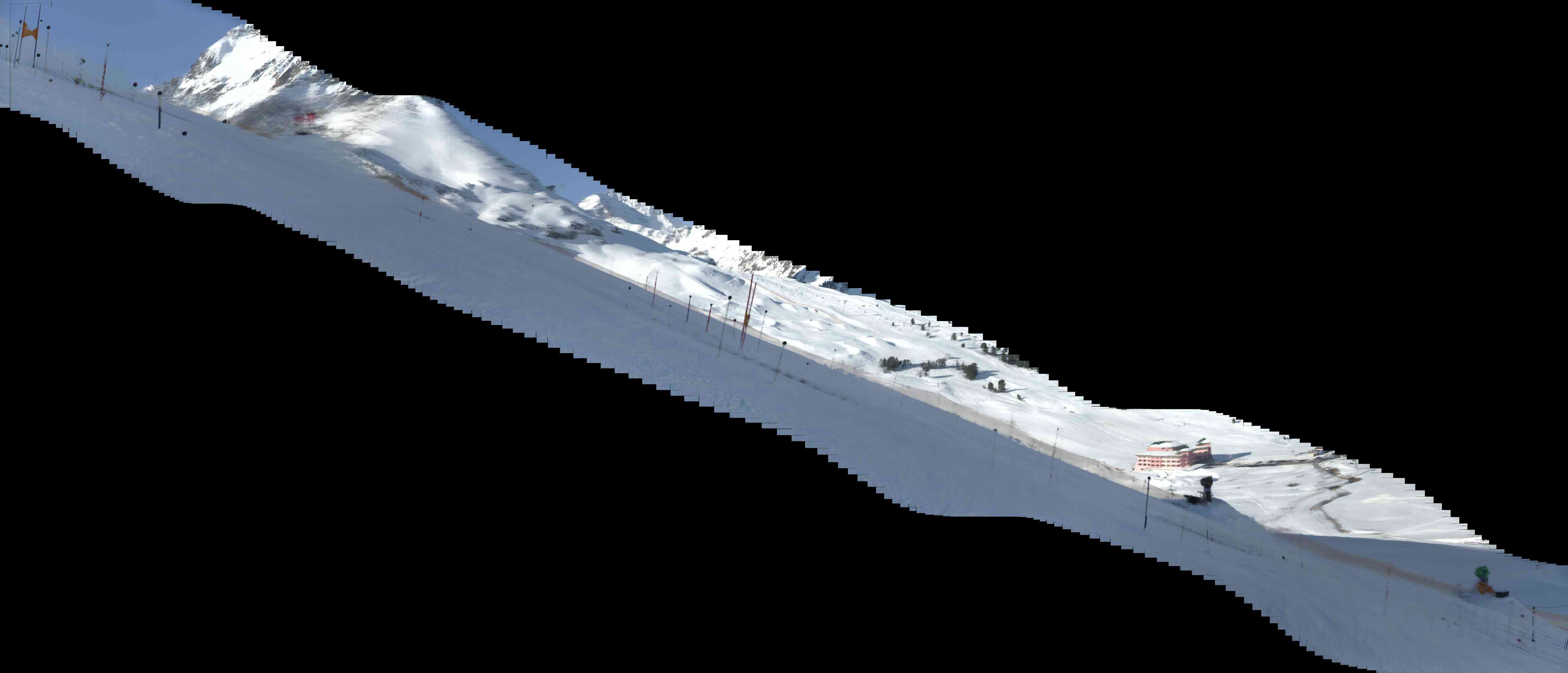}
	\caption{\small {\bf Ski dataset background samples.} Two samples of estimated ski backgrounds are shown.}
	\label{fig:ski_bg}
	\vspace{1cm}
\end{figure*}

In Section \ref{sec:SS-training}, we present training details and model architecture. In Section \ref{sec:qualitative}, we provide additional examples on disentanglement of pose and appearance, as well as qualitative results on pose estimation. Finally, we present details on gravity sensitivity analysis in Section \ref{sec:sup-gravity}.


\section{Training Procedure}
\label{sec:SS-training}

We use images of resolution $500 \times 500$ in the H36 and MPI datasets,  $500 \times 1000$ in the Diving dataset, and $640 \times 360$ in the Ski dataset. We train all models using the Adam optimizer~\cite{Kingma15} and a fixed learning rate of 1e-4. In theory, we could train our full model by simply minimizing the total loss $\mL_{\text{total}}$ of Eq.(\ref{eq:totalLoss}). However, because the model has to learn first where to detect the object, it takes a long time for the network to find the correct locations for the bounding boxes and convergence is slow. To speed it up, we introduce priors for the bounding box locations. 

For \textbf{H36M, MPI, and Ski}, we apply a Gaussian prior on the location and size of the bounding boxes. We set their average detected size over the mini-batch to be about 50\% of the input image resolution. On H36M we also set a prior that their average locations be in the middle of the frame. The latter was not needed on the MPI and Ski datasets. Note that these priors are applied to the average prediction of the mini-batch and not individual ones, which would otherwise imply a strong bias. We use this prior for all models including the baselines, except when spatial transformer networks are not used, as in the ablation study.

For \textbf{Diving}, which depicts free-fall motion, we use the gravity loss introduced in Section \ref{sec:gravity} in addition to the other objectives. It requires sampling frames that are temporally equidistant. Since this sampling is different from the one required for the DSL terms and the mini-batch size is limited, we developed a two-stage training strategy. First, the network is pre-trained with $\mL_{\text{DSL}}^{\text{all}}$ of Eq.~(\ref{eq:totalLoss}) with only gravity and reconstruction enabled, which helps localize the divers even though they occupy only a small fraction of image frame. Then, we switch on the DSL loss of $\mL_{\text{DSL}}^{\text{all}}$ of Eq.~(\ref{eq:totalLoss}) with its sampling to learn a more precise human model for which the gravity loss is no longer required. We use the gravity prior as pre-training for all models including the baselines because, without it,  it takes them longer to detect the subject. The only exception is Ours (no $\mathcal{L}_{track}$), which acts as an ablation study on the gravity prior. We use $\mathcal{L}_{track}$ only on this dataset and we set $\gamma$ to 1.

\textbf{Background Estimation.} As our model requires a background image, we apply the following procedure to estimate the background in each dataset. On H36M and MPI we use the per-pixel median, where the median is taken over all video frames. On the Diving dataset, we use the bottom half of the first frame and the top half of the last frame because the diver goes from the top of the frame to the bottom in the video. In the Ski dataset, the cameras rotate. We therefore compute a homography between consecutive frames using SIFT correspondences and generate a panorama in which we again take the median pixel values, as shown in Fig.~\ref{fig:ski_bg}.


\textbf{Hyper-parameters.} On the H36M, MPI, and Diving datasets we set $\lambda$ and $\rho$ to 2. On the Ski dataset, to help the model better capture the foreground, we reduce the RGB coefficient $\lambda$ to 0.05 while keeping $\rho$ at 2. This encourages the model to focus more on higher-level ImageNet features corresponding to structures in the image and hence to encode the skier better.
We set $\alpha$ to $0.01$ on the all datasets. We used the following temporal distance for sampling in models that exploit a contrastive loss: In H36M, a maximum of 10 temporal frame distance for positive samples and a minimum of 200 for negative samples. In MPI, 10 for positive and 450 for negative samples. In Diving and Ski, poses change more quickly within a few frames, so in Ski we use 5 for positive and 40 for negative samples whereas in Diving we use 3 for positive and 20 for negative samples. These hyper-parameters were selected based on how fast the poses change frame-to-frame and how the choice of these hyper-parameters impacts the quality of image reconstruction in the self-supervised models, which is our validation set model-selection criterion in the self-supervised model. We used the same sampling procedure for all models that use a contrastive loss.  

\subsection{Model Architecture}
\label{sec:impl}

To implement our model, we use a ResNet18 \cite{He16a} as detector that takes as input an image downsampled by factor $4$ in both dimensions and returns $\mS(\bI)=(s^x, s^y, u^x, u^y)$. This crops a patch $\bI_\text{crop}$ that is then resized to $128 \times 128$ using the $\text{affine-grid}$ PyTorch function. Unless otherwise specified, we use a ResNet50~\cite{He16a} model as the encoder for both $\mE_{ti}$ and $\mE_{tv}$. They share parameters, except in the output layer, where $\bI_\TI$ has a resolution of 129 and $\bI_\TV$ a resolution of 600 in H36M and Ski datasets and 900 in MPI and Dive datasets. In the decoder ${\mD}$, $\bI_\TV$ features are first passed to a fully-connected layer with ReLU and dropout with an output size of $32,768$, which is then reshaped to $128$ feature maps of resolution $16 \times 16$. The $\bI_\TI$ features are passed to another fully-connected layer with ReLU and dropout to produce an output of shape $128 \times 16^2$. The two feature maps are then concatenated into a single one of size $256 \times 16^2$. This feature map is then passed to three upsampling layers, each with one bilinear upsampling with factor 2 and three convolutional layers with kernel size $3 \times 3$. This upsamples the features maps from $256 \times 16^2$, in order, to $128 \times 32^2$, $64 \times 64^2$, and finally to $32 \times 128^2$. A final convolutional layer it then applied to make the resolution of these feature maps $4 \times 128^2$, where the first three channels are the decoded RGB image crop $\bD_{\text{crop}}$ and the last channel is the foreground subject mask crop $\bM_{\text{crop}}$. Finally, the inverse STN function is used to map the resolution of $\bD_{\text{crop}}$ and $\bM_{\text{crop}}$ to that of the original image and reconstructing the final image $\tilde{\bI}$ according to Eq.(\ref{eq:synthesize}). Figure \ref{fig:arch_details} depicts architecture details.

\textbf{Pose Model.}
	For the 3D pose estimation results presented in Sections \ref{sec:unsup_compare} and \ref{sec:analysis}, following \cite{Rhodin19a} we use a two-layer MLP for $\Phi$. It features two hidden layers of size 2048 with 50\% dropout applied to each hidden layer. The output layer dimension is $3 \times K$, where $K$ is the number of joints, which is equal to 17 in the H36M and Ski datasets, 24 for the 3DPW dataset, and to 13 in the Diving and MPI datasets. 
	
	For experiments in Section \ref{sec:exp_weak_labels}, unless otherwise specified, we use a multi-layer perceptron (MLP) ResNet architecture, taken from \cite{Martinez17a} for $\Phi$, which helps yielding more accurate 3D pose estimation. The ResNet consists of three residual blocks, each containing a two-layer MLP with ReLU non-linearity of dimension 1024. The three blocks is preceded by a linear layer that maps from time-variant resolution to 1024, and is followed by a linear layer that outputs the 3D pose. 
	
	When training the pose network, the self-supervised parameters are frozen, in order to evaluate the quality of extracted features in the self-supervised phase. The only except is the experiment in Figure~\ref{fig:imgNet_tv_comp}, where the impact of pre-training for model initialization is evaluated. For all models that split the latent features, we use only the time-variant component for pose estimation. For the models that do not---CSS, MV-CSS, AE-STN-Dec---we use all the latent features. 
	
	\textbf{Weak Labels.} When 2D pose label is used, we use the same architecture as our 3D MLP ResNet (without sharing parameters), with the difference that it outputs the 2D pose. The input to this model is the time-variant features. We set $\eta$ to 10 in all experiments that use 2D. 
	
	 Considering the upgraded encoder backbone in Section \ref{sec:exp_weak_labels}, we replace the ResNet50 with a ResNet101 module, taken from a pre-trained Alphapose model \cite{Fang17a} on CrowdPose \cite{Li18l} dataset. Note that this module is only part of the Alphapose model, hence we do not use its 2D predictions. The input image resolution to this model is $256 \times 320$ and it outputs a feature of dimentionality $2048 \times 10 \times 8$. To map it to our low-dimensional time-variant (tv) and time-invariant (ti) features, we use the same MLP residual block similar to the pose model described above, by changing its input to take ResNet101 features and its output to tv and ti dimensionality. 

\textbf{3D Pose Results.}  We report the results using the normalized mean per joint position error (N-MPJPE), expressed in mm, in Sections \ref{sec:unsup_compare} and \ref{sec:analysis}. The results in Section \ref{sec:exp_weak_labels} are reported with mean per joint position error (MPJPE), the normalized N-MPJPE, and the procrustes-aligned P-MPJPE.
 To test the performance of the models given only small sets of labeled data, we use different fractions of the training set. To this end, when the dataset is sub-sampled, we sub-sample every K frames of the training videos. Hence, sub-sampling is applied uniformly across dataset subjects and actions.

\subsection{Computational Complexity.}
A forward pass of the self-supervised network, including detector, encoder and decoder, for a mini-batch of size 32 takes 0.002 seconds on a a Tesla V100 GPU. We use the same backbone and model architecture for DrNet and NSD models. Thus, their computational complexity is almost the same as our model. By contrast, CSS and MV-CSS do not use the detection and decoding components. This makes training these models faster. However, as we showed  in Table~\ref{tab:h36}, these models are worse than our approach by a factor of around 2 and have a big gap compared to other models that reconstruct the image.

Considering the results in Section  \ref{sec:unsup_compare}, on H36M, the self-supervised models are trained for 250K iterations, which takes 4.5 days on a Tesla V100 GPU. The models converge at about 100K iterations, and further iterations yield negligible improvement. The pose models are trained for 50K iterations, which takes 4 hours on the same GPU.
On the MPI dataset, the self-supervised models are trained for 550K iterations and the pose models for 50K iterations.
On the Diving dataset, we trained self-supervised models for 500K iterations (taking about 7.25 days on a Tesla V100 GPU). Due to GPU limitations we trained on mini-batches of size 16 for this dataset. We used mini-batched of size 32 on all other datasets. The pose models were trained for 40K iterations, taking about 9 hours on the same GPU.  
On the Ski dataset, the self-supervised models are trained for 400K (taking about 4 days on Tesla V100 GPU) and the pose models for 100K iterations. These same settings were applied to all models trained on a given dataset.

Considering the results in Section \ref{sec:exp_weak_labels}, on the H36M and MPI datasets, the self-supervised models are trained for 200K and the pose models for 100K iterations. On the Diving dataset, the self-supervised models are trained for 300K and the pose models for 200K iterations.

\section{Qualitative Results} 
\label{sec:qualitative}
\subsection{Disentanglement.} In Figures \ref{fig:supp_swap1} we take time-invariant component from an image and time-variant component from a different image to reconstruct an output image. As can be observed, the model learns to separate these two components. 

\subsection{Prediction Samples.}
\begin{figure*}[h]
	\centering
	\includegraphics[width=1\textwidth]{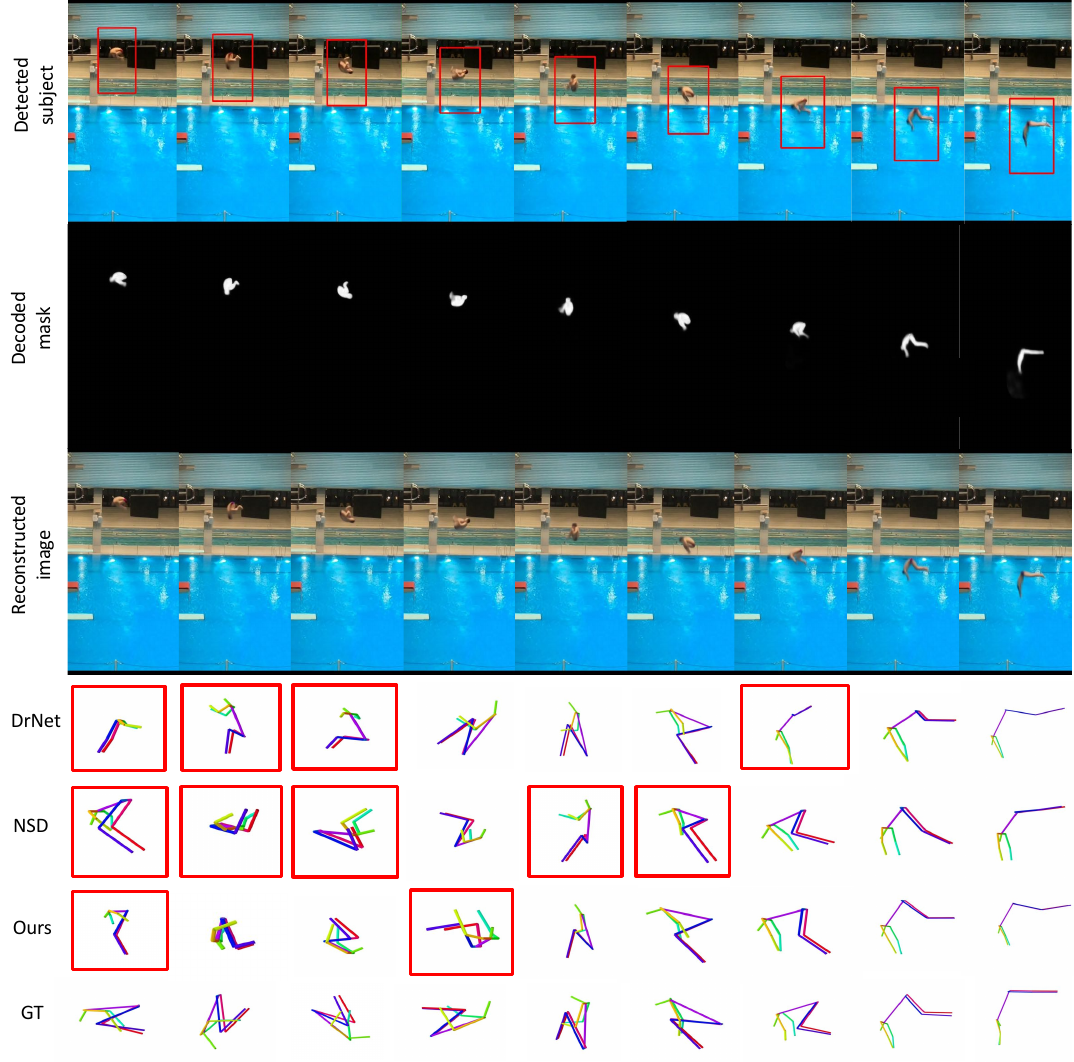}
	\caption{\small {\bf Qualitative results on Diving Dataset}.  Samples are taken from the test set. In the top row, the red  rectangle shows the bounding box detected by our model for image cropping. The second and third rows, show respectively the decoded mask and image. Bottom rows, show the pose estimation by different models, where the red rectangle indicates pose estimation with a high error. GT in the bottom row indicates the ground truth labels.}
	\label{fig:dive_pose1}
\end{figure*}

\begin{figure*}[t]
	\centering
	\includegraphics[width=1\textwidth]{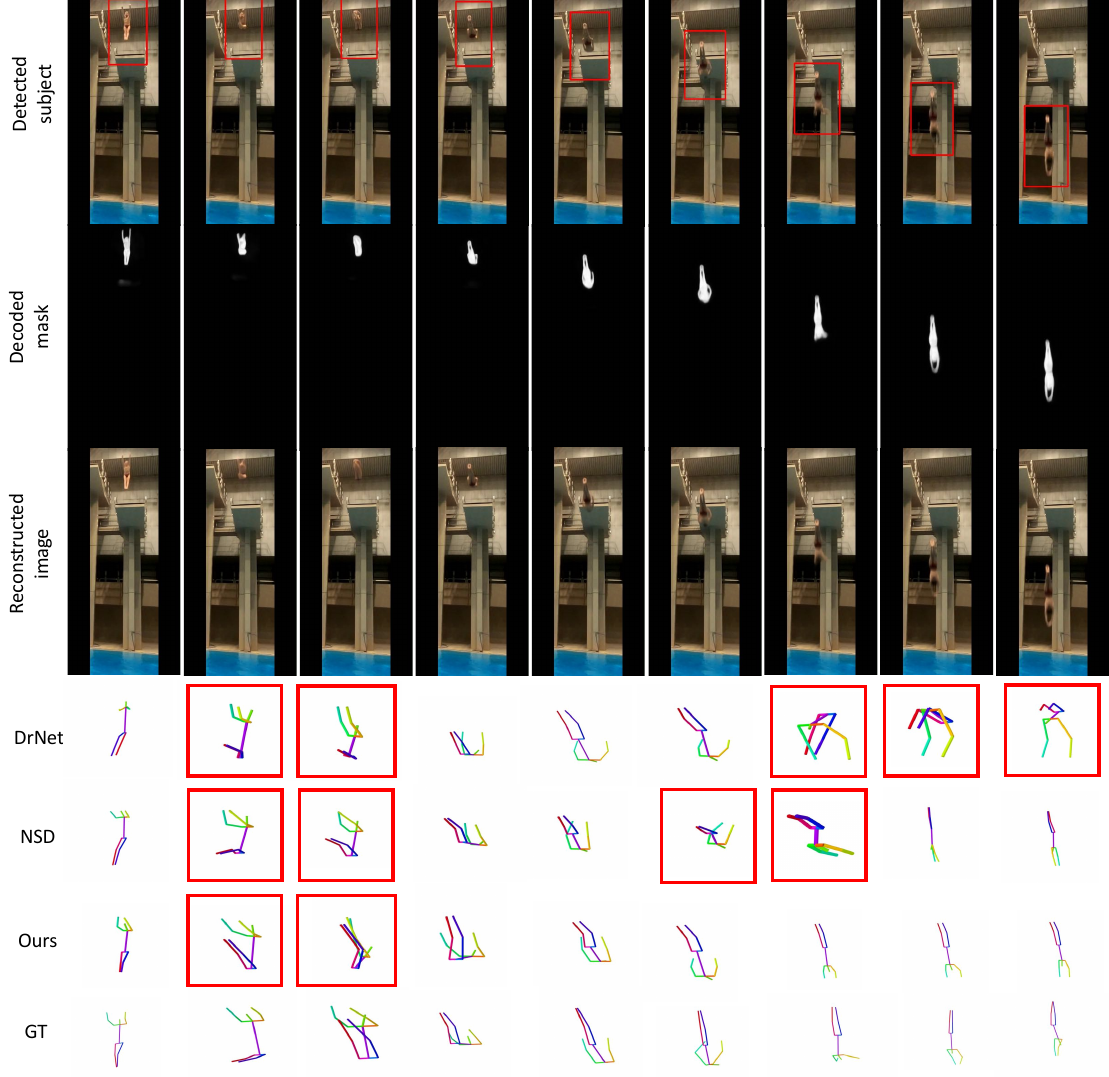}
	\caption{\small {\bf Qualitative results on Diving Dataset}.  Samples are taken from the test set. In the top row, the red  rectangle shows the bounding box detected by our model for image cropping. The second and third rows, show respectively the decoded mask and image. Bottom rows, show the pose estimation by different models, where the red rectangle indicates pose estimation with a high error. GT in the bottom row indicates the ground truth labels.}
	\label{fig:supp_dive_pose2}
\end{figure*}

\begin{figure*}[t]
	\centering
	\includegraphics[width=1\textwidth]{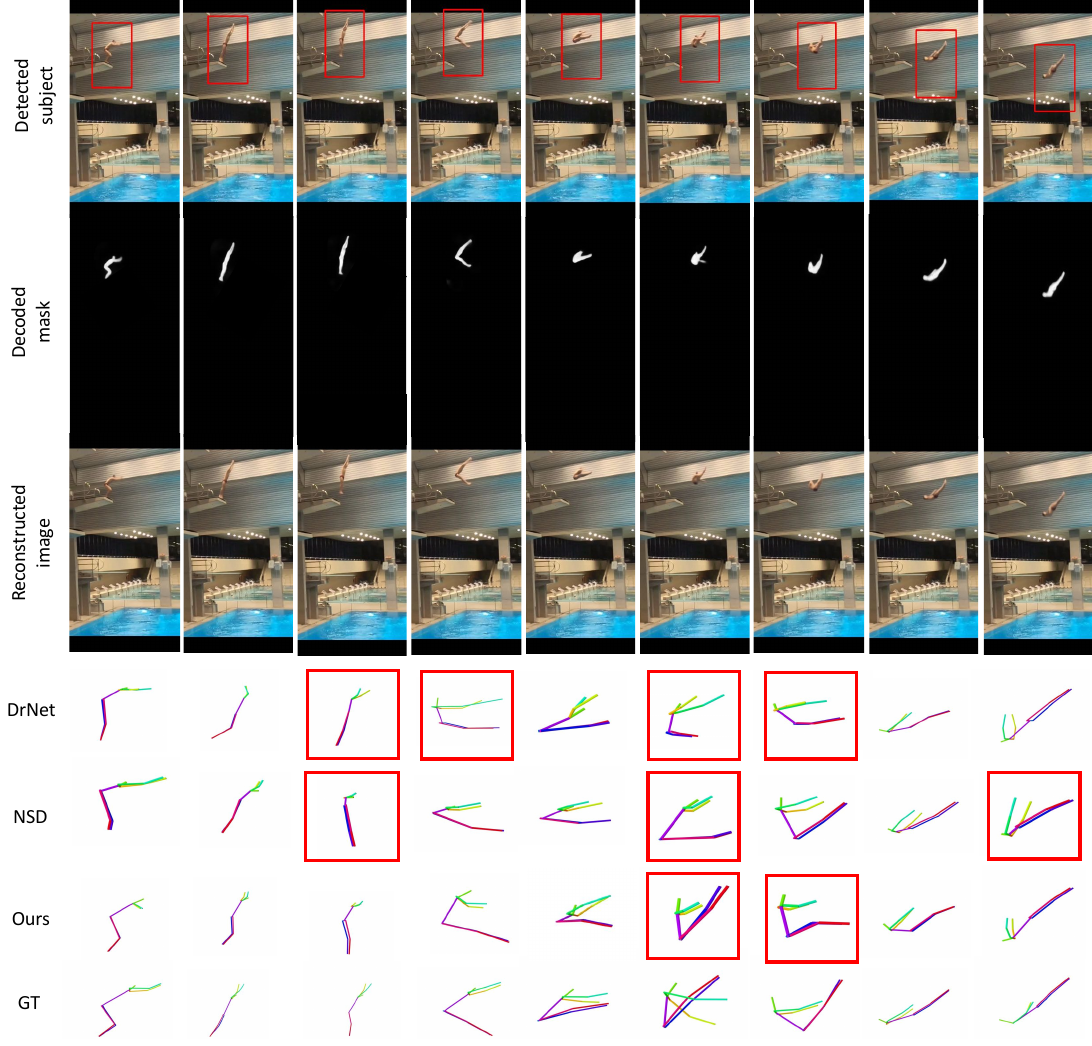}
	\caption{\small {\bf Qualitative results on Diving Dataset}.  Samples are taken from the test set. In the top row, the red  rectangle shows the bounding box detected by our model for image cropping. The second and third rows, show respectively the decoded mask and image. Bottom rows, show the pose estimation by different models, where the red rectangle indicates pose estimation with a high error. GT in the bottom row indicates the ground truth labels.}
	\label{fig:supp_dive_pose3}
	\vspace{1cm}
\end{figure*}

\begin{figure*}[h!]
	\centering
	\begin{subfigure}[b]{0.15\textwidth}
	\includegraphics[width=\textwidth]{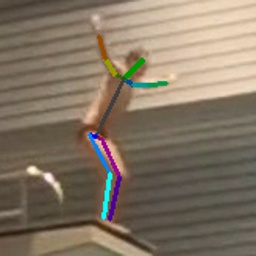} 
	\end{subfigure}
	\hspace{0.01em}
	\begin{subfigure}[b]{0.15\textwidth}
		\includegraphics[width=\textwidth]{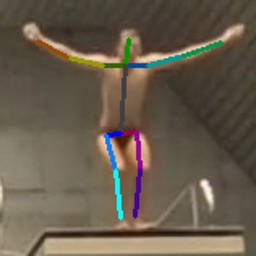} 
	\end{subfigure}
	\hspace{0.01em}
	\begin{subfigure}[b]{0.15\textwidth}
	\includegraphics[width=\textwidth]{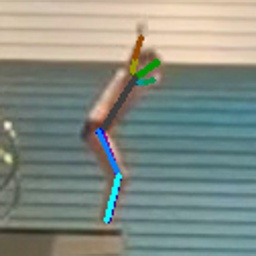}
	\end{subfigure}
	\hspace{0.2cm}
	\begin{subfigure}[b]{0.15\textwidth}
		\includegraphics[width=\textwidth]{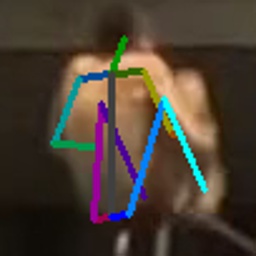} 
	\end{subfigure}
	\hspace{0.01em}
	\begin{subfigure}[b]{0.15\textwidth}
		\includegraphics[width=\textwidth]{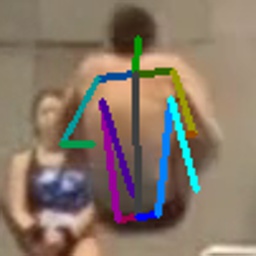}
	\end{subfigure}
	\hspace{0.01em}
	\begin{subfigure}[b]{0.15\textwidth}
		\includegraphics[width=\textwidth]{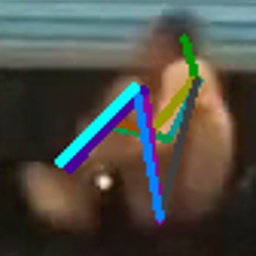}
	\end{subfigure}
	
	 \vspace{0.2cm }
	 
	 	\centering
	 \begin{subfigure}[b]{0.15\textwidth}
	 	\includegraphics[width=\textwidth]{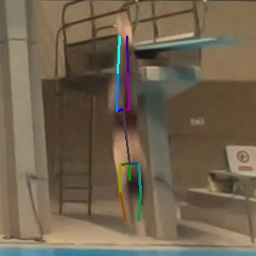} 
	 \end{subfigure}
	 \hspace{0.01em}
	 \begin{subfigure}[b]{0.15\textwidth}
	 	\includegraphics[width=\textwidth]{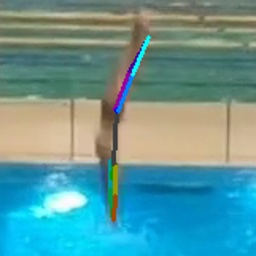} 
	 \end{subfigure}
	 \hspace{0.01em}
	 \begin{subfigure}[b]{0.15\textwidth}
	 	\includegraphics[width=\textwidth]{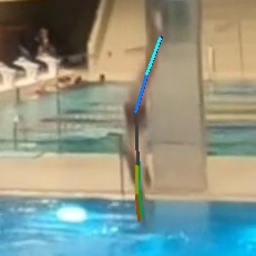}
	 \end{subfigure}
	 \hspace{0.2cm}
	 \begin{subfigure}[b]{0.15\textwidth}
	 	\includegraphics[width=\textwidth]{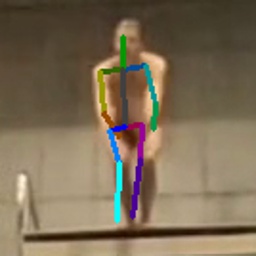} 
	 \end{subfigure}
	 \hspace{0.01em}
	 \begin{subfigure}[b]{0.15\textwidth}
	 	\includegraphics[width=\textwidth]{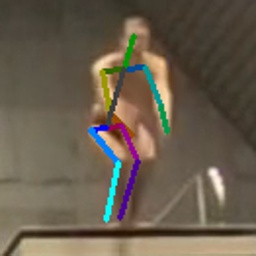}
	 \end{subfigure}
	 \hspace{0.01em}
	 \begin{subfigure}[b]{0.15\textwidth}
	 	\includegraphics[width=\textwidth]{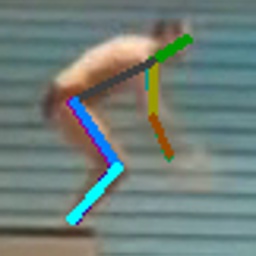}
	 \end{subfigure}
	 
	 \vspace{0.2cm }
	 
	 \centering
	 \begin{subfigure}[b]{0.15\textwidth}
	 	\includegraphics[width=\textwidth]{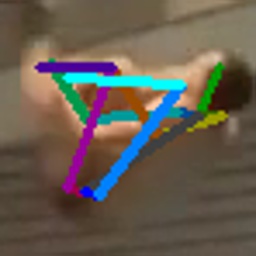} 
	 \end{subfigure}
	 \hspace{0.01em}
	 \begin{subfigure}[b]{0.15\textwidth}
	 	\includegraphics[width=\textwidth]{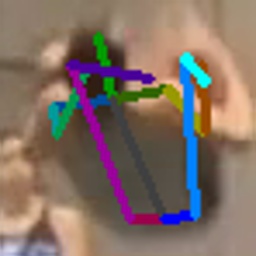} 
	 \end{subfigure}
	 \hspace{0.01em}
	 \begin{subfigure}[b]{0.15\textwidth}
	 	\includegraphics[width=\textwidth]{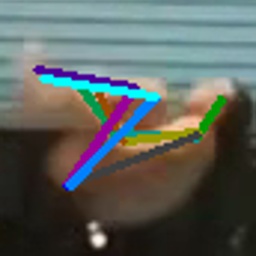}
	 \end{subfigure}
	 \hspace{0.2cm}
	 \begin{subfigure}[b]{0.15\textwidth}
	 	\includegraphics[width=\textwidth]{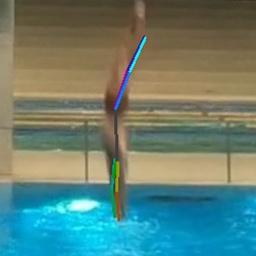} 
	 \end{subfigure}
	 \hspace{0.01em}
	 \begin{subfigure}[b]{0.15\textwidth}
	 	\includegraphics[width=\textwidth]{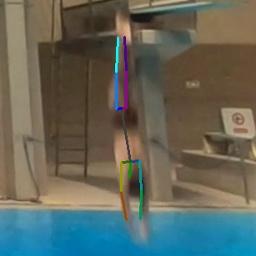}
	 \end{subfigure}
	 \hspace{0.01em}
	 \begin{subfigure}[b]{0.15\textwidth}
	 	\includegraphics[width=\textwidth]{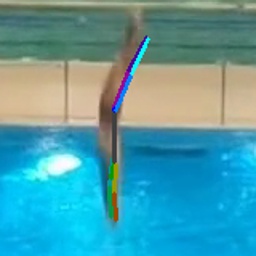}
	 \end{subfigure}

	\vspace{0.2cm }

	\centering
	\begin{subfigure}[b]{0.15\textwidth}
		\includegraphics[width=\textwidth]{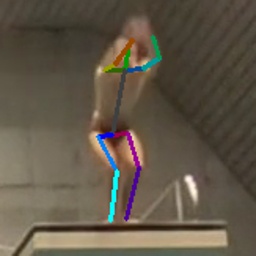} 
	\end{subfigure}
	\hspace{0.01em}
	\begin{subfigure}[b]{0.15\textwidth}
		\includegraphics[width=\textwidth]{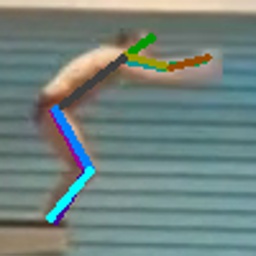} 
	\end{subfigure}
	\hspace{0.01em}
	\begin{subfigure}[b]{0.15\textwidth}
		\includegraphics[width=\textwidth]{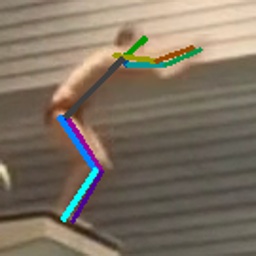}
	\end{subfigure}
	\hspace{0.2cm}
	\begin{subfigure}[b]{0.15\textwidth}
		\includegraphics[width=\textwidth]{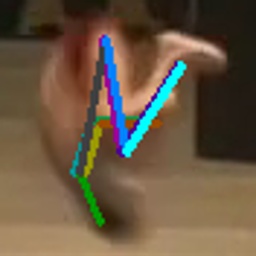} 
	\end{subfigure}
	\hspace{0.01em}
	\begin{subfigure}[b]{0.15\textwidth}
		\includegraphics[width=\textwidth]{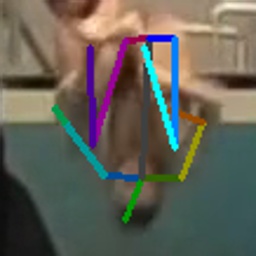}
	\end{subfigure}
	\hspace{0.01em}
	\begin{subfigure}[b]{0.15\textwidth}
		\includegraphics[width=\textwidth]{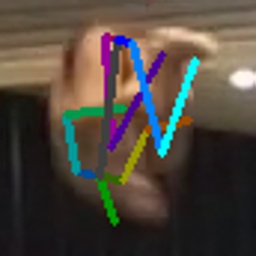}
	\end{subfigure}
	 
	 \vspace{0.2cm }
	
	\centering
	\begin{subfigure}[b]{0.15\textwidth}
		\includegraphics[width=\textwidth]{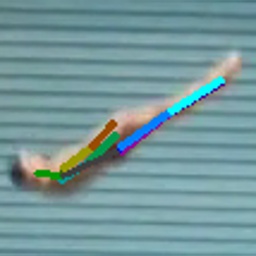} 
	\end{subfigure}
	\hspace{0.01em}
	\begin{subfigure}[b]{0.15\textwidth}
		\includegraphics[width=\textwidth]{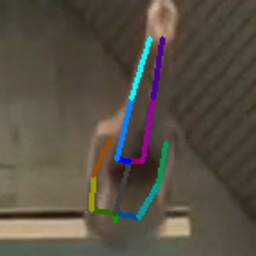} 
	\end{subfigure}
	\hspace{0.01em}
	\begin{subfigure}[b]{0.15\textwidth}
		\includegraphics[width=\textwidth]{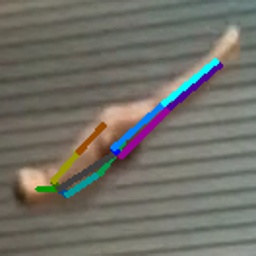}
	\end{subfigure}
	\hspace{0.2cm}
	\begin{subfigure}[b]{0.15\textwidth}
		\includegraphics[width=\textwidth]{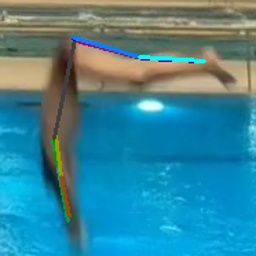} 
	\end{subfigure}
	\hspace{0.01em}
	\begin{subfigure}[b]{0.15\textwidth}
		\includegraphics[width=\textwidth]{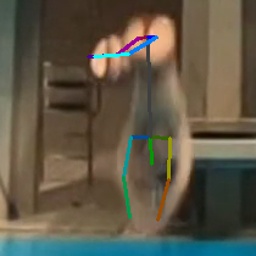} 
	\end{subfigure}
	\hspace{0.01em}
	\begin{subfigure}[b]{0.15\textwidth}
		\includegraphics[width=\textwidth]{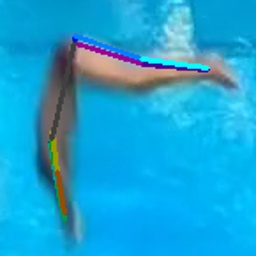}
	\end{subfigure}
		 
	\vspace{0.2cm }
	
	\centering
	\begin{subfigure}[b]{0.15\textwidth}
		\includegraphics[width=\textwidth]{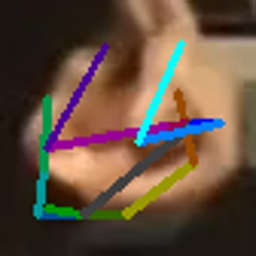} 
	\end{subfigure}
	\hspace{0.01em}
	\begin{subfigure}[b]{0.15\textwidth}
		\includegraphics[width=\textwidth]{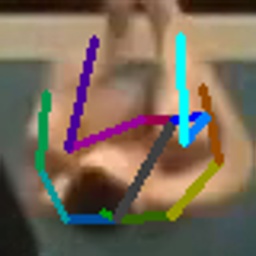} 
	\end{subfigure}
	\hspace{0.01em}
	\begin{subfigure}[b]{0.15\textwidth}
		\includegraphics[width=\textwidth]{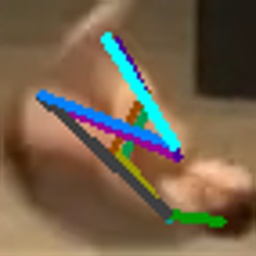}
	\end{subfigure}
	\hspace{0.2cm}
	\begin{subfigure}[b]{0.15\textwidth}
		\includegraphics[width=\textwidth]{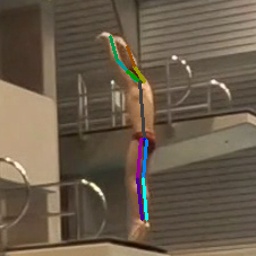} 
	\end{subfigure}
	\hspace{0.01em}
	\begin{subfigure}[b]{0.15\textwidth}
		\includegraphics[width=\textwidth]{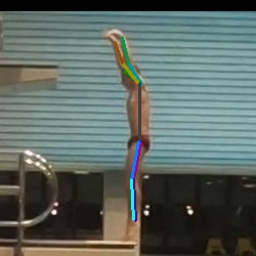}
	\end{subfigure}
	\hspace{0.01em}
	\begin{subfigure}[b]{0.15\textwidth}
		\includegraphics[width=\textwidth]{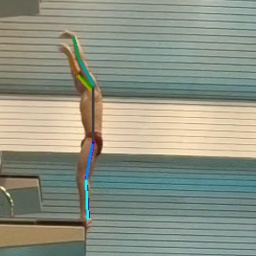}
	\end{subfigure}

	 \vspace{-0.1cm }
	 \centering \rule{17.5cm}{0.01cm}
	\vspace{0.2cm }
	
	\centering
	\begin{subfigure}[b]{0.15\textwidth}
		\includegraphics[width=\textwidth]{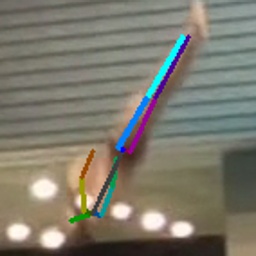} \caption*{Original View}
	\end{subfigure}
	\hspace{0.01em}
	\begin{subfigure}[b]{0.15\textwidth}
		\includegraphics[width=\textwidth]{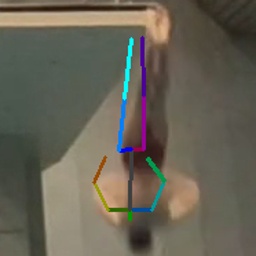} \caption*{Side-View 1}
	\end{subfigure}
	\hspace{0.01em}
	\begin{subfigure}[b]{0.15\textwidth}
		\includegraphics[width=\textwidth]{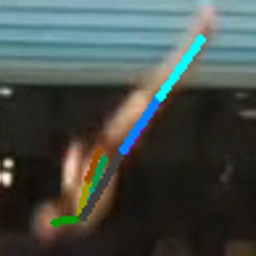} \caption*{Side-View 2}
	\end{subfigure}
	\hspace{0.2cm}
	\begin{subfigure}[b]{0.15\textwidth}
		\includegraphics[width=\textwidth]{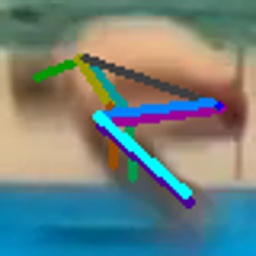} \caption*{Original View}
	\end{subfigure}
	\hspace{0.01em}
	\begin{subfigure}[b]{0.15\textwidth}
		\includegraphics[width=\textwidth]{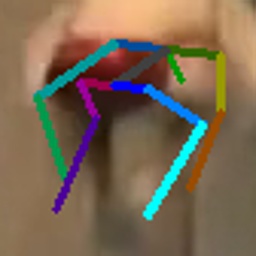} \caption*{Side-View 1}
	\end{subfigure}
	\hspace{0.01em}
	\begin{subfigure}[b]{0.15\textwidth}
		\includegraphics[width=\textwidth]{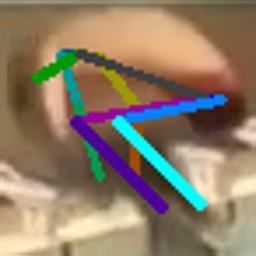} \caption*{Side-View 2}	
	\end{subfigure}
	
	\caption{\small 
		Side-view visualization on diving dataset. For each triplet, the first image shows the prediction of the model on the input view, while side-views 1 and 2 show projection of the 3D pose to other views. The last row shows erroneous cases where the model's prediction in the original view is plausible while it does not correspond to correct poses in other views.
	}
	\label{fig:side_view_dive}
\end{figure*}

\begin{figure*}[h!]
	\centering
	\begin{subfigure}[b]{0.15\textwidth}
	\includegraphics[width=\textwidth]{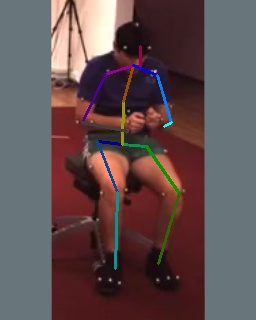} 
	\end{subfigure}
	\hspace{0.01em}
	\begin{subfigure}[b]{0.15\textwidth}
		\includegraphics[width=\textwidth]{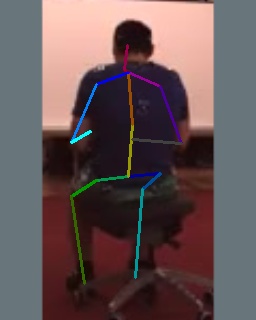} 
	\end{subfigure}
	\hspace{0.01em}
	\begin{subfigure}[b]{0.15\textwidth}
	\includegraphics[width=\textwidth]{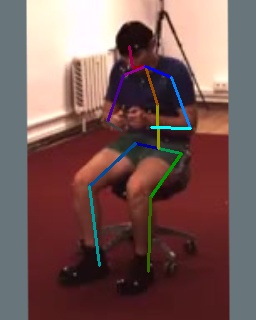}
	\end{subfigure}
	\vspace{0.2cm}
	\begin{subfigure}[b]{0.15\textwidth}
		\includegraphics[width=\textwidth]{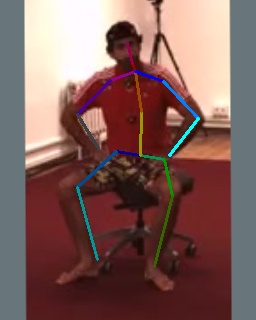} 
	\end{subfigure}
	\hspace{0.01em}
	\begin{subfigure}[b]{0.15\textwidth}
		\includegraphics[width=\textwidth]{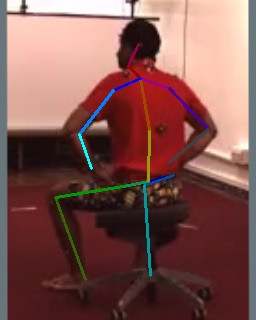}
	\end{subfigure}
	\hspace{0.01em}
	\begin{subfigure}[b]{0.15\textwidth}
		\includegraphics[width=\textwidth]{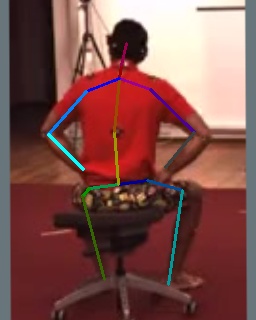}
	\end{subfigure}
	
	 \vspace{0.2cm }
	 
	 \centering
	 \begin{subfigure}[b]{0.15\textwidth}
	 	\includegraphics[width=\textwidth]{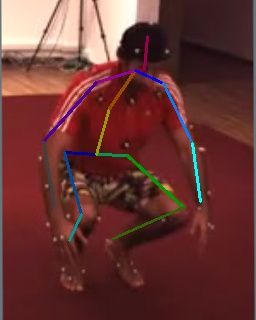} 
	 \end{subfigure}
	 \hspace{0.01em}
	 \begin{subfigure}[b]{0.15\textwidth}
	 	\includegraphics[width=\textwidth]{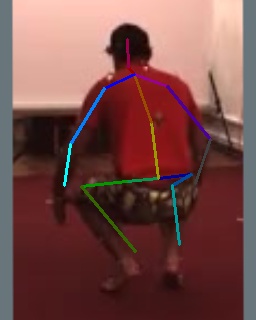} 
	 \end{subfigure}
	 \hspace{0.01em}
	 \begin{subfigure}[b]{0.15\textwidth}
	 	\includegraphics[width=\textwidth]{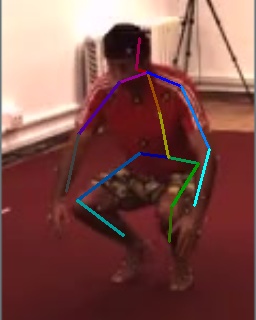}
	 \end{subfigure}
	 \hspace{0.2cm}
	 \begin{subfigure}[b]{0.15\textwidth}
	 	\includegraphics[width=\textwidth]{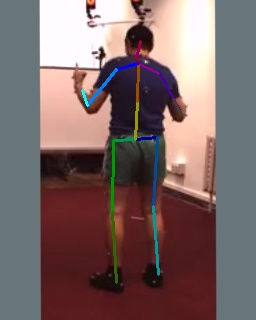} 
	 \end{subfigure}
	 \hspace{0.01em}
	 \begin{subfigure}[b]{0.15\textwidth}
	 	\includegraphics[width=\textwidth]{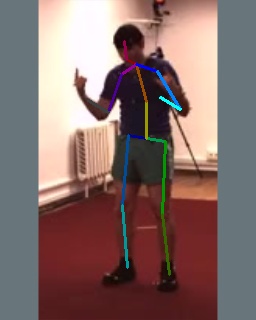}
	 \end{subfigure}
	 \hspace{0.01em}
	 \begin{subfigure}[b]{0.15\textwidth}
	 	\includegraphics[width=\textwidth]{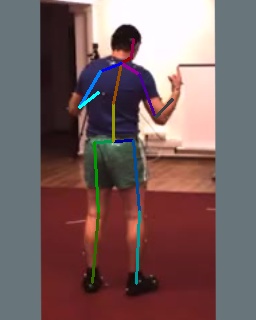}
	 \end{subfigure}
	 \vspace{0.2cm }
	 
	 \centering
	 \begin{subfigure}[b]{0.15\textwidth}
	 	\includegraphics[width=\textwidth]{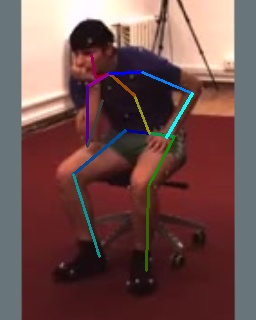} 
	 \end{subfigure}
	 \hspace{0.01em}
	 \begin{subfigure}[b]{0.15\textwidth}
	 	\includegraphics[width=\textwidth]{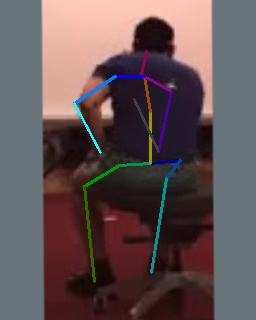}
	 \end{subfigure}
	 \hspace{0.01em}
	 \begin{subfigure}[b]{0.15\textwidth}
	 	\includegraphics[width=\textwidth]{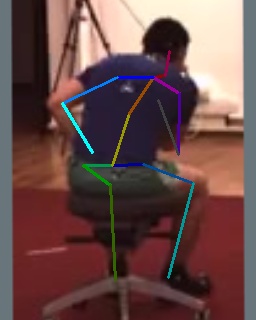}
	 \end{subfigure}
	 \hspace{0.2cm}
	 \begin{subfigure}[b]{0.15\textwidth}
	 	\includegraphics[width=\textwidth]{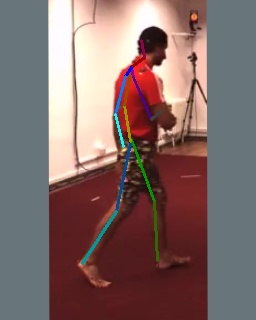} 
	 \end{subfigure}
	 \hspace{0.01em}
	 \begin{subfigure}[b]{0.15\textwidth}
	 	\includegraphics[width=\textwidth]{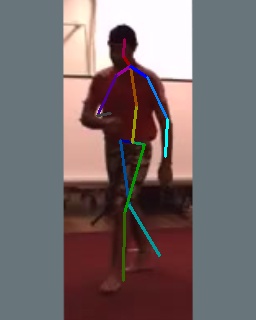} 
	 \end{subfigure}
	 \hspace{0.01em}
	 \begin{subfigure}[b]{0.15\textwidth}
	 	\includegraphics[width=\textwidth]{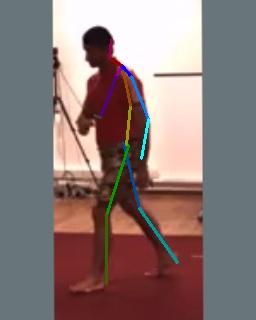}
	 \end{subfigure}
	 
	 \vspace{0.2cm }
	
	\centering
	\begin{subfigure}[b]{0.15\textwidth}
		\includegraphics[width=\textwidth]{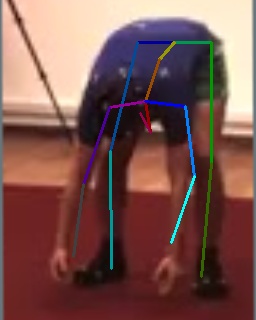}
	\end{subfigure}
	\hspace{0.01em}
	\begin{subfigure}[b]{0.15\textwidth}
		\includegraphics[width=\textwidth]{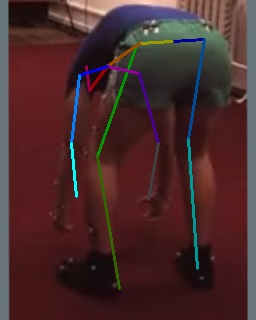}
	\end{subfigure}
	\hspace{0.01em}
	\begin{subfigure}[b]{0.15\textwidth}
		\includegraphics[width=\textwidth]{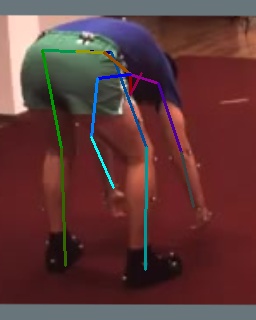}
	\end{subfigure}
	\hspace{0.2cm}
	\begin{subfigure}[b]{0.15\textwidth}
		\includegraphics[width=\textwidth]{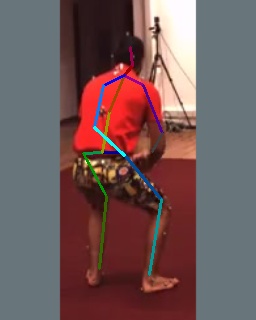}
	\end{subfigure}
	\hspace{0.01em}
	\begin{subfigure}[b]{0.15\textwidth}
		\includegraphics[width=\textwidth]{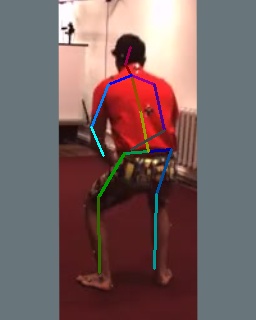}
	\end{subfigure}
	\hspace{0.01em}
	\begin{subfigure}[b]{0.15\textwidth}
		\includegraphics[width=\textwidth]{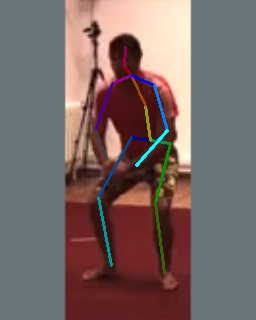}
	\end{subfigure}

	 \vspace{0.2cm }

	\centering
	\begin{subfigure}[b]{0.15\textwidth}
		\includegraphics[width=\textwidth]{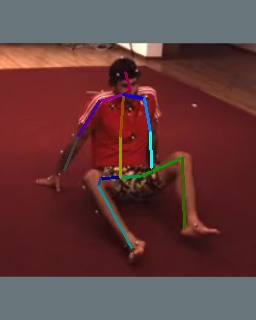}
	\end{subfigure}
	\hspace{0.01em}
	\begin{subfigure}[b]{0.15\textwidth}
		\includegraphics[width=\textwidth]{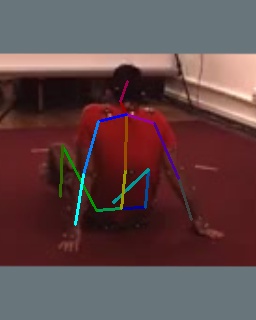}
	\end{subfigure}
	\hspace{0.01em}
	\begin{subfigure}[b]{0.15\textwidth}
		\includegraphics[width=\textwidth]{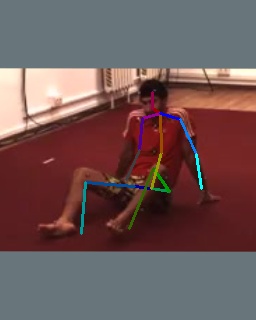}
	\end{subfigure}
	\hspace{0.2cm}
	\begin{subfigure}[b]{0.15\textwidth}
		\includegraphics[width=\textwidth]{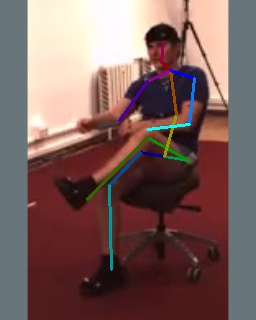}
	\end{subfigure}
	\hspace{0.01em}
	\begin{subfigure}[b]{0.15\textwidth}
		\includegraphics[width=\textwidth]{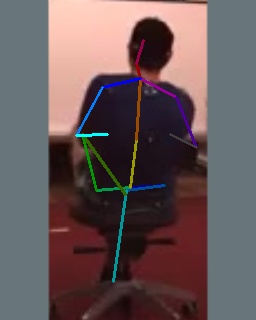}
	\end{subfigure}
	\hspace{0.01em}
	\begin{subfigure}[b]{0.15\textwidth}
		\includegraphics[width=\textwidth]{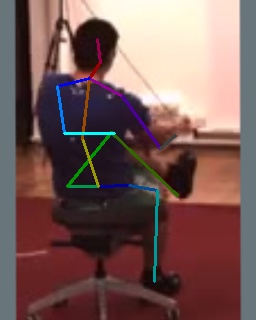}
	\end{subfigure}
		
	\vspace{-0.1cm }
	\centering \rule{17.5cm}{0.01cm}
	\vspace{0.2cm }
	
	\centering
	\begin{subfigure}[b]{0.15\textwidth}
		\includegraphics[width=\textwidth]{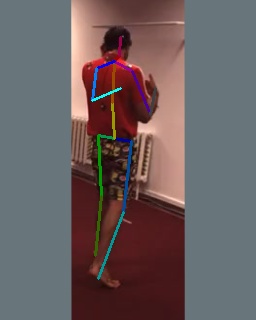} \caption*{Original View}
	\end{subfigure}
	\hspace{0.01em}
	\begin{subfigure}[b]{0.15\textwidth}
		\includegraphics[width=\textwidth]{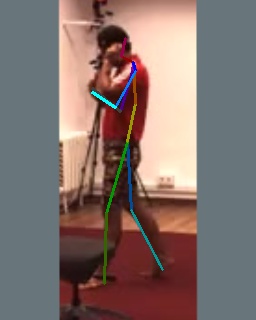} \caption*{Side-View 1}
	\end{subfigure}
	\hspace{0.01em}
	\begin{subfigure}[b]{0.15\textwidth}
		\includegraphics[width=\textwidth]{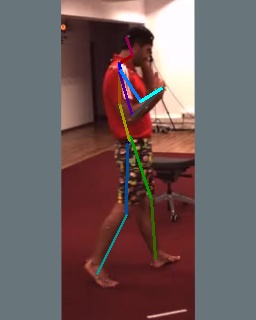} \caption*{Side-View 2}
	\end{subfigure}
	\hspace{0.2cm}
	\begin{subfigure}[b]{0.15\textwidth}
		\includegraphics[width=\textwidth]{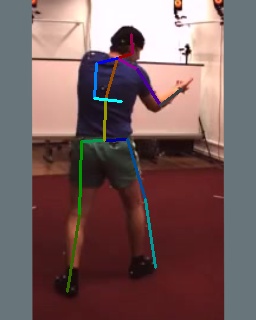} \caption*{Original View}
	\end{subfigure}
	\hspace{0.01em}
	\begin{subfigure}[b]{0.15\textwidth}
		\includegraphics[width=\textwidth]{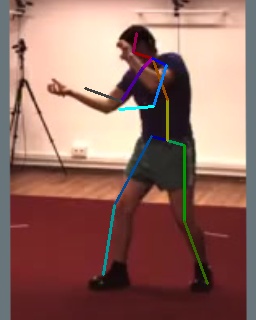} \caption*{Side-View 1}
	\end{subfigure}
	\hspace{0.01em}
	\begin{subfigure}[b]{0.15\textwidth}
		\includegraphics[width=\textwidth]{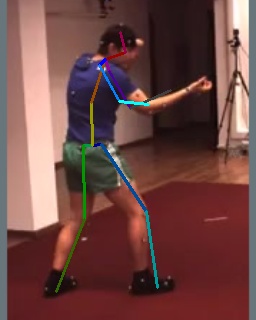} \caption*{Side-View 2}
	\end{subfigure}
	
	\caption{\small 
			Side-view visualization on H36M dataset. For each triplet, the first image shows the prediction of the model on the input view, while side-views 1 and 2 show projection of the 3D pose to other views. The last row shows erroneous cases where the model's prediction in the original view is plausible while it does not correspond to correct poses in other views.
	}
	\label{fig:side_view_h36}
\end{figure*}

In Figures \ref{fig:supp_mpi_pose} and \ref{fig:supp_posetrack}  we show pose estimation results on MPI and H36M test sets. In Figures \ref{fig:dive_pose1},
 \ref{fig:supp_dive_pose2}, \ref{fig:supp_dive_pose3}, 
 we show samples on the Diving dataset and visualize detected bounding box, segmentation mask, decoded image, as well as final pose predictions by different models.  We also visualize side-view results on H36M and Diving datasets in Figures \ref{fig:side_view_dive}  and \ref{fig:side_view_h36} to show the quality of the learned depth.
 
Following \cite{Bieler19}, we measure the center of gravity (CG) of the body by using the ground truth 3D joints, where we first get the CG location of each body-part (such as forearm, upper-arm, thigh), as provided in \cite{Clauser71}, then we measure the CG of the entire body as the weighted average of CG of each body-part multiplied in its weight ratio (as provided in \cite{Clauser71}), yielding $\text{CG}_\text{body} = \sum_j^J w_j \text{CG}_j$, where $w_j$ is the weight ratio of body part $j$ w.r.t to the entire body weight and $\text{CG}_j$ is the 3D location of the CG of body part $j$. We then project the estimated CG to the image using camera projection matrix and also visualize our center of bounding box (bbox), which estimates the CG. While the estimated CG and center of bbox have a gap, our estimated center of bbox can guide the model to correctly localize the person.



\begin{figure*}[htbp]
	\centering
	\includegraphics[width=1\textwidth]{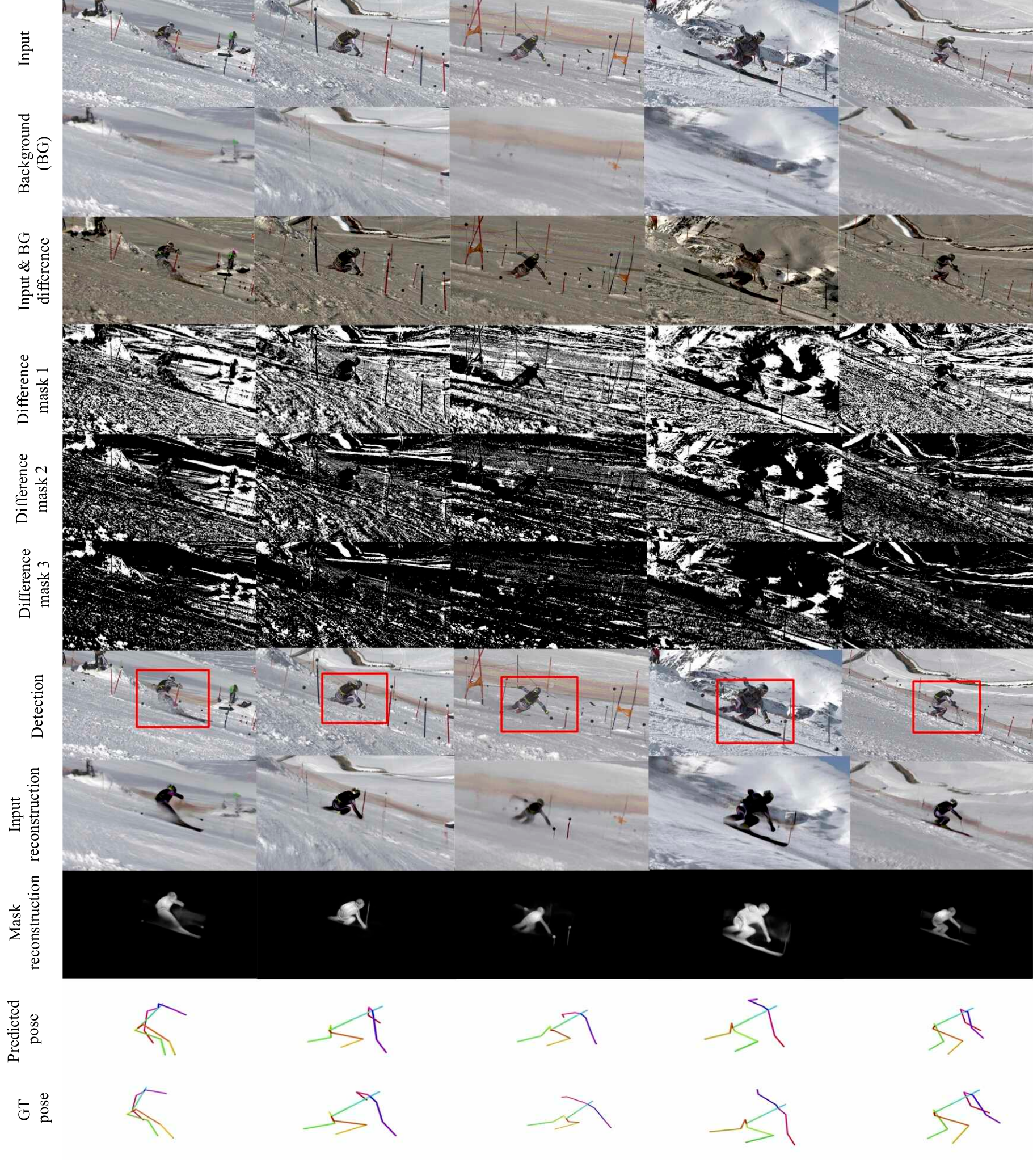}
	\caption{\small {\bf Ski dataset test samples.} In the top two rows we show the input images and our estimated backgrounds. In the third row we show the background subtracted input image. As can be observed the background is still present in the image. We also apply thresholding to this image and create masks by using three different thresholds. The resulted images are shown in rows 4 to 6. The thresholding still does not remove the background and even increasing the threshold does not help. In the next four rows we show the output of our model, which are detected subject bounding box, the reconstructed image, the predicted mask, and finally the estimated pose.}
	\label{fig:ski_big}
\end{figure*}


\section{Acceleration Sensitivity Analysis}
\label{sec:sup-gravity}
An object's location $\bp(t)$ at time $t$ falling down in 3D space using gravity is formulated by $\bp(t)=0.5 \bg t^2 + \bv_0 t+ \bp_0$, where $\bg$, $\bv_0$, and $\bp_0$ respectively indicating gravity, initial speed,and  initial location.
We can write the projection of the center of gravity $CG$ in the image plane in projective coordinates as
\begin{align}
	\bp_{\text{proj}} = [p_{\text{proj}}^x, p_{\text{proj}}^y, p_{\text{proj}}^z] = \bK [\bR | \bT] (0.5 \bg t^2 + \bv_0 t+ \bp_0),
\end{align}
where $\bg$, $\bv_0$, and $\bp_0$ are represented in homogeneous coordinates, each being a column vector of $4 \times 1$.
this yields
\begin{align}
	\begin{bmatrix} 
		p_{\text{proj}}^x \\
		p_{\text{proj}}^y  \\
		p_{\text{proj}}^z
	\end{bmatrix} &= 
	\begin{bmatrix} 
		\bK_1^T [\bR | \bT] (0.5 \bg t^2 + \bv_0 t+ \bp_0) \\
		\bK_2 ^T[\bR | \bT] (0.5 \bg t^2 + \bv_0 t+ \bp_0) \\
		\bK_3^T [\bR | \bT] (0.5 \bg t^2 + \bv_0 t+ \bp_0) 
	\end{bmatrix} 
\end{align}
where $\bK_i^T$ is the $i$-th row of matrix $\bK$.
The observed $x$ and $y$ locations are then equal to
\begin{align}
	p^x =  p_{\text{proj}}^x  / p_{\text{proj}}^z, \; \; \;
	p^y =  p_{\text{proj}}^y / p_{\text{proj}}^z \; . \nonumber
\end{align}

When the image plane is parallel to the plane in which the subject travels or in orthographic projection, the depth $p_{\text{proj}}^z$ remains fixed or is ignored in the course of dive, so we can treat is as a constant $c$. In this case, we then get 
\begin{align}
	\begin{bmatrix} 
		p^x \\
		p^y 
	\end{bmatrix} &= 
	\frac{1}{c}
	\begin{bmatrix} 
		\bK_1^T [\bR | \bT] (0.5 \bg t^2 + \bv_0 t+ \bp_0) \\
		\bK_2 ^T[\bR | \bT] (0.5 \bg t^2 + \bv_0 t+ \bp_0) 
	\end{bmatrix} 
\end{align}

Getting the second order derivative yields
\begin{align}
	\begin{bmatrix} 
		a^x \\
		a^y 
	\end{bmatrix} &= 
	\frac{1}{c}
	\begin{bmatrix} 
		\bK_1^T [\bR | \bT] \bg \\
		\bK_2 ^T[\bR | \bT] \bg 
	\end{bmatrix} 
\end{align}

with $a^x $ and $a^y$  being acceleration in $x$ and $y$ directions. This makes acceleration a constant value dependent on intrinsic $\bK$, rotation $\bR$, translation $\bT$, and gravity $\bg$, hence not a function of time $t$. In this case, our assumption of constant acceleration strictly holds. 

On the other hand, if the subject  travels on a plane not parallel to the image plane, the depth of observed location changes over time, making it a function of $t$. In this case for $y$ location we have ($p^x$ is obtained similarly)
\begin{align}
	p^y = \frac{p_{\text{proj}}^y}{p_{\text{proj}}^z}=  \frac{\bK_2^T [\bR | \bT] (0.5 \bg t^2 + \bv_0 t+ \bp_0)}{\bK_3^T [\bR | \bT] (0.5 \bg t^2 + \bv_0 t+ \bp_0)}
\end{align}

Getting derivative w.r.t $t$ we have
\begin{align}
	& v^y = \frac{\partial p^y }{\partial t}=  \\ 
	& \frac{\bK_2^T [\bR | \bT](\bg t+ \bv_0)(p_{\text{proj}}^z) - \bK_3^T [\bR | \bT](\bg t+ \bv_0)(p_{\text{proj}}^y)}{(p_{\text{proj}}^z)^2} = \nonumber \\
	&\frac{ (p_{\text{proj}}^z \bK_2^T - p_{\text{proj}}^y\bK_3^T) ([\bR | \bT](\bg t+ \bv_0))}{(p_{\text{proj}}^z)^2} \nonumber
\end{align}
Getting second derivative w.r.t $t$ we obtain
\begin{align}
	& a^y = \frac{\partial v^y }{\partial t}= \label{eq:acceleration}\\ 
	&\frac{ (p_{\text{proj}}^z \bK_2^T - p_{\text{proj}}^y\bK_3^T) ([\bR | \bT]\bg)(p_{\text{proj}}^z)^2}{(p_{\text{proj}}^z)^4} - \nonumber \\
	&\frac{2 p_{\text{proj}}^z  (\frac{\partial }{\partial t}p_{\text{proj}}^z)(p_{\text{proj}}^z \bK_2^T - p_{\text{proj}}^y\bK_3^T) ([\bR | \bT](\bg t+ \bv_0))}{(p_{\text{proj}}^z)^4} = \nonumber \\
	& \text{\footnotesize $\frac{p_{\text{proj}}^z(p_{\text{proj}}^z \bK_2^T - p_{\text{proj}}^y\bK_3^T) [\bR | \bT](p_{\text{proj}}^z \bg  - (2 (\frac{\partial }{\partial t}p_{\text{proj}}^z)(\bg t+ \bv_0)))}{(p_{\text{proj}}^z)^4} \nonumber$}
\end{align}
As can be seen in the last line of Eq.(\ref{eq:acceleration}),  both the nominator and denominator depend on $p_{\text{proj}}^z$, which is the depth. It changes over time and is therefore a function of $t$. This violates our assumption that the acceleration is constant in the image plane when the subject moves in a plane not parallel to the image plane. However, $p_{\text{proj}}^z$ only changes slowly for points far away from the camera because perspective projection gets increasingly close to orthographic projection as one moves away from the camera.

\end{document}